\documentclass[10pt,journal,compsoc]{IEEEtran}

\usepackage{pgfplots}
\pgfplotsset{compat=newest}
\usepackage{tikz}

\ifCLASSOPTIONcompsoc
  \usepackage[nocompress]{cite}
\else
  \usepackage{cite}
\fi

\usepackage{amsmath}
\usepackage{amssymb}
\usepackage{amsthm}
\usepackage{mathtools}

\ifCLASSOPTIONcompsoc
 \usepackage[caption=false,font=footnotesize,labelfont=sf,textfont=sf]{subfig}
\else
 \usepackage[caption=false,font=footnotesize]{subfig}
\fi

\usepackage{times}
\usepackage{epsfig}
\usepackage{graphicx}

\usepackage{color}
\usepackage{xcolor}
\usepackage{booktabs,siunitx} 
\usepackage{makecell}
\usepackage{enumitem}
\usepackage[percent]{overpic}
\usepackage[caption=false]{subfig}
\usepackage{rotating}
\usepackage[misc]{ifsym}
\usepackage{multirow}

\usepackage[pagebackref=true,breaklinks=true,colorlinks,bookmarks=false]{hyperref}

\if@mathematic
   
\else
   
\fi

\newcommand{\blue}[1]{\textcolor{black}{#1}}

\definecolor{road}                {RGB}{128, 64,128}
\definecolor{sidewalk}            {RGB}{244, 35,232}
\definecolor{building}            {RGB}{ 70, 70, 70}
\definecolor{wall}                {RGB}{102,102,156}
\definecolor{fence}               {RGB}{190,153,153}
\definecolor{pole}                {RGB}{153,153,153}
\definecolor{traffic light}       {RGB}{250,170, 30}
\definecolor{traffic sign}        {RGB}{220,220,  0}
\definecolor{vegetation}          {RGB}{107,142, 35}
\definecolor{terrain}             {RGB}{152,251,152}
\definecolor{sky}                 {RGB}{ 70,130,180}
\definecolor{person}              {RGB}{220, 20, 60}
\definecolor{rider}               {RGB}{255,  0,  0}
\definecolor{car}                 {RGB}{  0,  0,142}
\definecolor{truck}               {RGB}{  0,  0, 70}
\definecolor{bus}                 {RGB}{  0, 60,100}
\definecolor{train}               {RGB}{  0, 80,100}
\definecolor{motorcycle}          {RGB}{  0,  0,230}
\definecolor{bicycle}             {RGB}{119, 11, 32}
\definecolor{void}                {RGB}{  0,  0,  0}

\newcommand\ver[1]{\rotatebox[origin=c]{90}{#1}}

\begin{document}

\title{Condition-Invariant Semantic Segmentation}

\author{Christos~Sakaridis,
        David~Bruggemann,
        Fisher~Yu,
        and~Luc~Van~Gool
\IEEEcompsocitemizethanks{\IEEEcompsocthanksitem C.~Sakaridis is with the Department of Information Technology and Electrical Engineering, ETH Z\"urich, Switzerland.\protect\\
E-mail: csakarid@vision.ee.ethz.ch
\IEEEcompsocthanksitem D.~Bruggemann is with the Department of Information Technology and Electrical Engineering, ETH Z\"urich, Switzerland.
\IEEEcompsocthanksitem F.~Yu is with the Department of Information Technology and Electrical Engineering, ETH Z\"urich, Switzerland.
\IEEEcompsocthanksitem L.~Van Gool is with the Department of Information Technology and Electrical Engineering, ETH Z\"urich, Switzerland, with the Department of Electrical Engineering, KU Leuven, Belgium, and with INSAIT, Bulgaria.}
}

\markboth{IEEE Transactions on Pattern Analysis and Machine Intelligence,~Vol.~xx, No.~xx, July~2024}%
{Sakaridis \MakeLowercase{\textit{et al.}}: Condition-Invariant Semantic Segmentation}

\IEEEtitleabstractindextext{%
\begin{abstract}
Adaptation of semantic segmentation networks to different visual conditions is vital for robust perception in autonomous cars and robots. However, previous work has shown that most feature-level adaptation methods, which employ adversarial training and are validated on synthetic-to-real adaptation, provide marginal gains in condition-level adaptation, being outperformed by simple pixel-level adaptation via stylization. Motivated by these findings, we propose to leverage stylization in performing feature-level adaptation by aligning the internal network features extracted by the encoder of the network from the original and the stylized view of each input image with a novel feature invariance loss. In this way, we encourage the encoder to extract features that are already invariant to the style of the input, allowing the decoder to focus on parsing these features and not on further abstracting from the specific style of the input. We implement our method, named Condition-Invariant Semantic Segmentation (CISS), on the current state-of-the-art domain adaptation architecture and achieve outstanding results on condition-level adaptation. In particular, CISS sets the new state of the art in the popular daytime-to-nighttime Cityscapes$\to$Dark Zurich benchmark. Furthermore, our method achieves the second-best performance on the normal-to-adverse Cityscapes$\to$ACDC benchmark. CISS is shown to generalize well to domains unseen during training, such as BDD100K-night \blue{and ACDC-night}. Code is publicly available at \url{https://github.com/SysCV/CISS}.
\end{abstract}

}

\maketitle

\IEEEdisplaynontitleabstractindextext

\IEEEpeerreviewmaketitle

\IEEEraisesectionheading{\section{Introduction}\label{sec:intro}}

\IEEEPARstart{U}{nsupervised} domain adaptation (UDA) is a primary instance of transfer learning, in which a labeled source set and an unlabeled target set are given at training time and the goal is to optimize performance on the domain of the latter set. There is a large body of literature focusing on UDA for semantic segmentation, which is of high practical importance for central computer vision applications such as autonomous cars and robots, as these systems need to have a dense pixel-level parsing of their surrounding scene, are bound to encounter data from different domains than those annotated for training, and labeling large quantities of data for each new deployment domain is very time- and cost-intensive. The main directions of recent research on this task are adversarial learning for domain alignment~\cite{adapt:structured:output:cvpr18,advent:adaptation,bidirectional:learning:adaptation,category:adversaries:adaptation,sim:adaptation} and training with pseudolabels~\cite{self:training:adaptation,crst:adaptation,mrnet:rectifying,dacs:domain:adaptation,diga:adaptation,IR2FRMM:adaptation}, with methods primarily focusing on the synthetic-to-real UDA setting~\cite{playing:data,Synthia:dataset}, i.e., GTA5$\to$Cityscapes and SYNTHIA$\to$Cityscapes. However, the normal-to-adverse Cityscapes$\to$ACDC UDA benchmark introduced in~\cite{ACDC} showed that adversarial-learning-based methods, which attempt to align domains at the level of features, struggle with the domain shift from normal to adverse conditions. By contrast, Fourier domain adaptation (FDA)~\cite{fda:adaptation} was shown in~\cite{ACDC} to provide significant gains in this normal-to-adverse setting, even with its simple non-learned pixel-level domain alignment.

\begin{figure}
  \centering
  \subfloat[Image]{\includegraphics[width=0.495\linewidth]{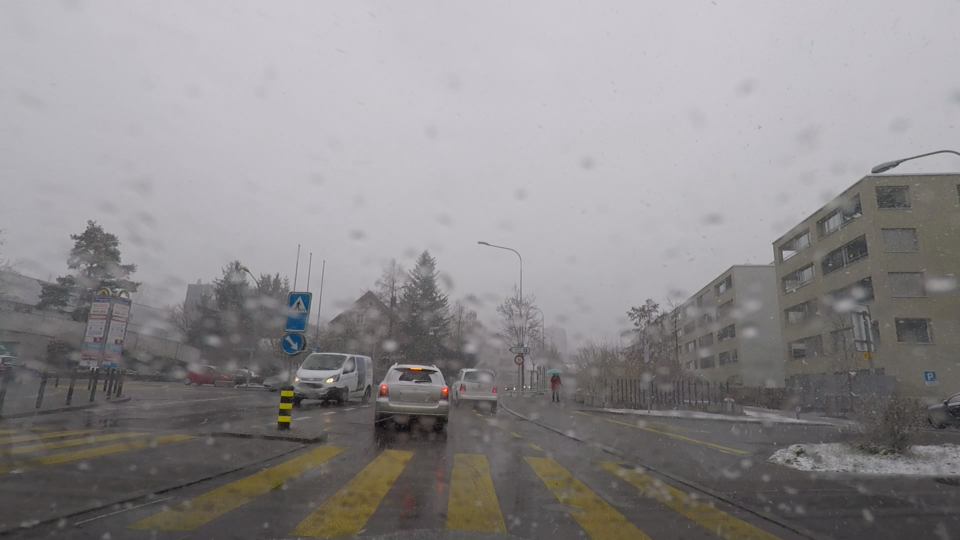}}
  \hfil
  \subfloat[Ground-truth semantics]{\includegraphics[width=0.495\linewidth]{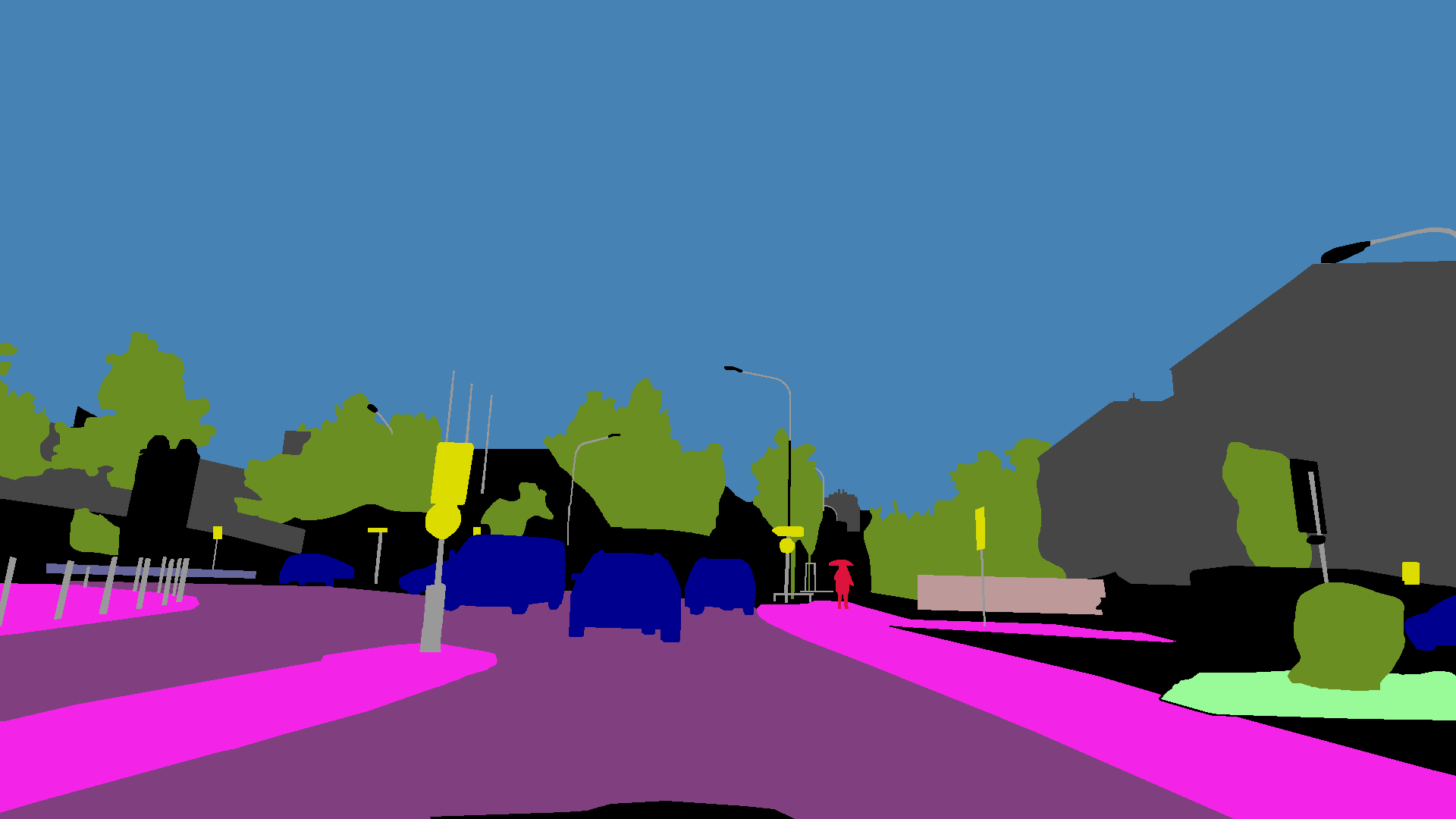}}
  \\
  \vspace{-0.35cm}
  \subfloat[HRDA~\cite{hrda:domain:adaptation}]{\includegraphics[width=0.495\linewidth]{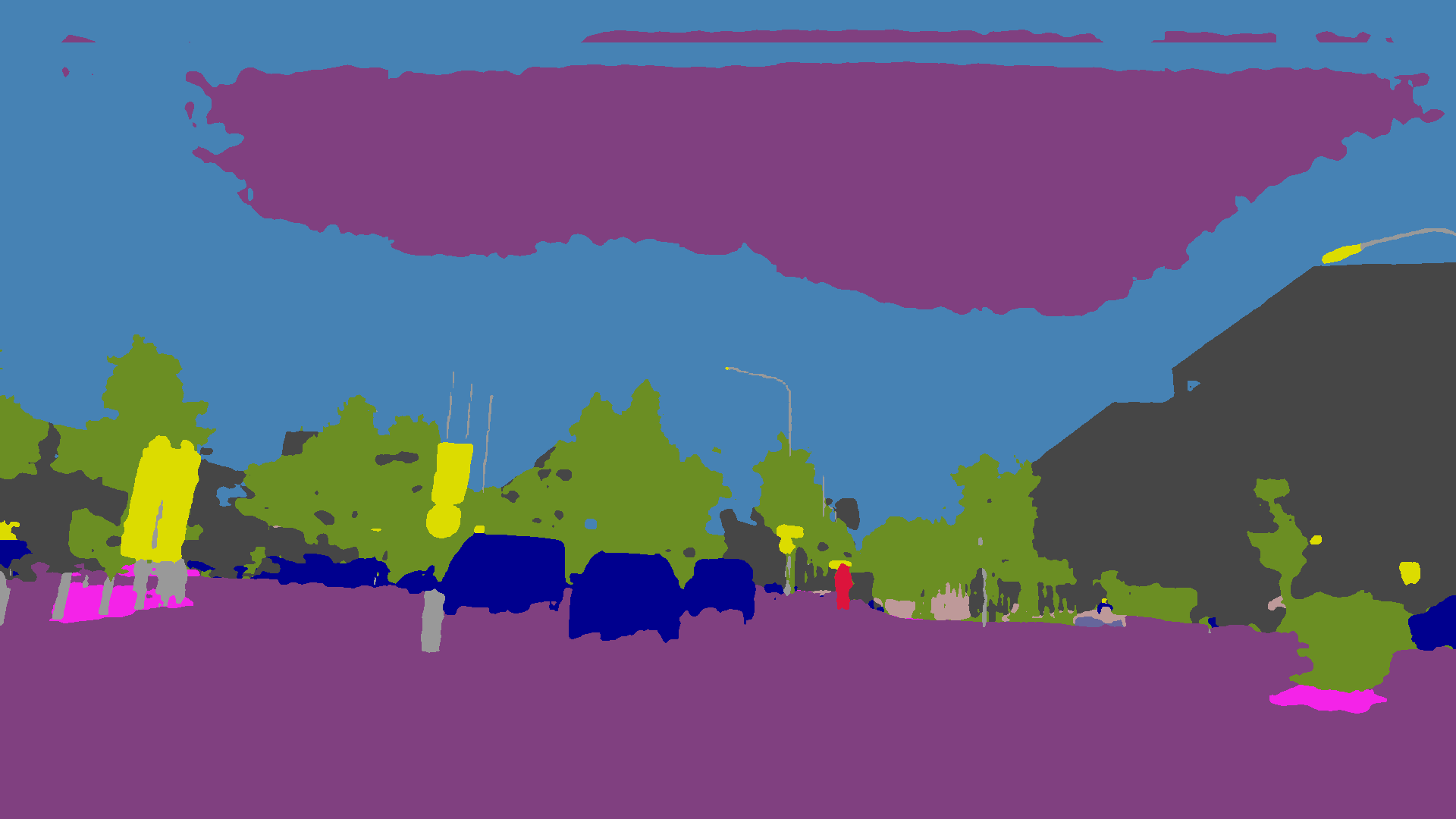}\label{fig:teaser:hrda}}
  \hfil
  \subfloat[CISS (ours)]{\includegraphics[width=0.495\linewidth]{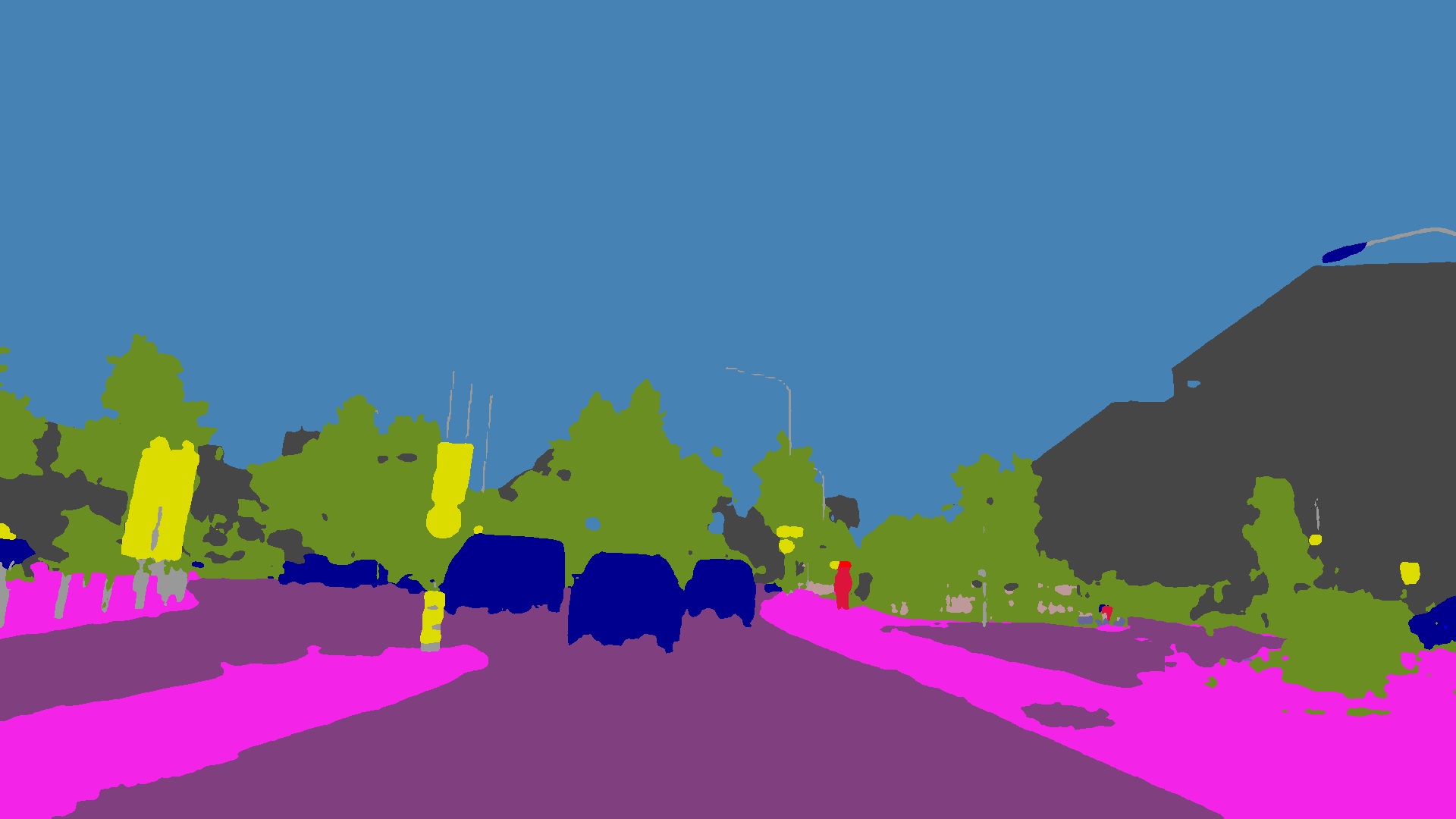}}
  \vspace{+0.1cm}
  \caption{The domain shift from normal to adverse conditions presents challenges to top-performing state-of-the-art domain adaptation methods for semantic segmentation \protect\subref{fig:teaser:hrda} due to the large resulting change in the appearance of classes. We propose a method that encourages invariance of internal features of segmentation networks to visual conditions by comparing features of different views of the same scene under the style of different domains, improving segmentation especially for classes which undergo large shifts.}
  \label{fig:teaser}
\end{figure}

We recognize that the problem with adversarial approaches is that they discriminate between feature maps that are extracted from \emph{different} scenes, which does not allow to disentangle the difference in the domain from the difference in the scene content. The key idea in this work is to factor out the aforementioned difference in scene content by aligning \emph{internal} features which are extracted from two versions of the \emph{same} scene that belong to different domains with a feature invariance loss that penalizes differences between the two feature maps. The intuition is that the encoder of the semantic segmentation network should output features that are already \emph{invariant} to the domain/style of the scene, so that the decoder can subsequently produce identical outputs for the different versions of the scene, as the ground-truth semantics of these versions are also identical. To our knowledge, we are the first to propose this cross-domain internal feature invariance in UDA for semantic segmentation, which hinges on comparing features from different views of the same scene rendered in different domains/styles, \blue{and we demonstrate through our experiments the superiority of our \emph{internal} feature invariance to the \emph{output-level} consistency which is invariably employed in the literature.}

A major challenge in implementing the novel feature invariance loss is the generation of representative alternative views of input source-domain or target-domain scenes. Instead of relying on learned models which add significant complexity to the overall adaptation architecture or on simple photometric augmentations, we propose to leverage shallow stylization methods, e.g.\ FDA~\cite{fda:adaptation} or simple color transfer~\cite{reinhard2001color}, to this end. In order to transfer each source-domain image to the style of the target domain, we use the corresponding target-domain image of the training mini-batch and transfer its style to the source-domain image. This allows a light-weight stylization that is simply implemented as part of the data loading in training. The original and stylized source-domain images are then both fed to the segmentation network to compute the feature invariance loss. The converse procedure is followed for each target-domain image of each training mini-batch. As the invariance of features is promoted across views of the scene which are characterized by an identical structure of the objects that are present, we term our method Condition-Invariant Semantic Segmentation (CISS, pronounced \emph{kiss}). The name of our method signifies that it is tailored for condition-level domain shifts and not shifts involving structural changes of objects, as in the synthetic-to-real setup, where the shape of objects may change across domains. CISS is not specific to the particular stylization it uses and works well with different stylization techniques including \cite{fda:adaptation,reinhard2001color}, as we evidence in Sec.~\ref{sec:experiments}.

In our experiments, we use the state-of-the-art HRDA~\cite{hrda:domain:adaptation} architecture and implement CISS on top of it. We show that our feature invariance loss improves significantly upon the straightforward alternative of defining an extra cross-entropy loss on the stylized images and we demonstrate the merit of applying this loss to internal features instead of output-level representations, contrary to previous works. Moreover, the separate feature invariance losses on source and target images are shown to be synergistic, leading to state-of-the-art results both on the Cityscapes$\to$Dark Zurich and Cityscapes$\to$ACDC UDA benchmarks. More specifically, on Cityscapes$\to$Dark Zurich, CISS not only outperforms HRDA by 4.8\% in mean IoU, but it also beats MIC~\cite{mic:adaptation}, which is the previous best-performing method on this benchmark and also builds on top of HRDA in a direction orthogonal to CISS. Our method is additionally ranked second on Cityscapes$\to$ACDC, delivering a significant improvement over HRDA and performing competitively to MIC across all four adverse conditions of the test set of ACDC, where it achieves the top performance on the rain split. Last but not least, we evaluate the CISS model trained for nighttime segmentation on Cityscapes$\to$Dark Zurich in zero-shot generalization setting\blue{s} using the BDD100K-night set~\cite{MGCDA_UIoU,BDD100K} \blue{and the ACDC-night set~\cite{ACDC}} and demonstrate the benefit of condition invariance for generalization across diverse unseen nighttime data.

\begin{figure*}
    \centering
    \includegraphics[clip,width=0.95\textwidth,trim=48mm 19mm 54mm 19mm]{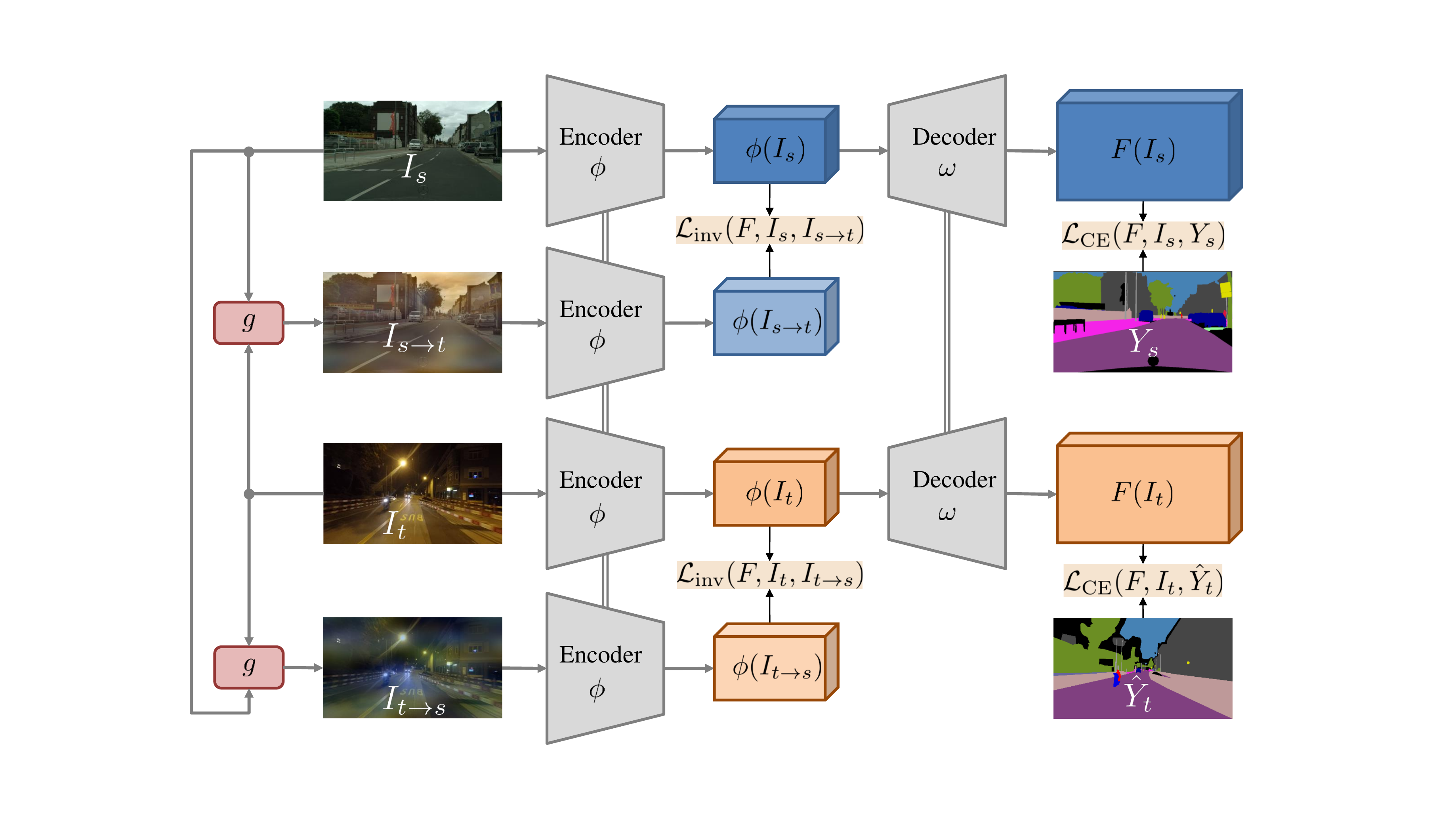}
    \caption{\textbf{Overview of our method.} Two instances of a shallow stylization mapping $g$ are fed with the source and target image, $I_s$ and $I_t$, to produce versions stylized with the converse domain, $I_{s\to{}t}$ and $I_{t\to{}s}$. \blue{In this example, $I_{s\to{}t}$ and $I_{t\to{}s}$ are computed using FDA~\cite{fda:adaptation}.} The four images are fed to a shared encoder $\phi$, the features of which are used to compute our feature invariance losses. The features of the original source and target images are further fed to a shared decoder $\omega$ to compute softmax predictions and respective cross-entropy losses. Double lines indicate shared weights.}
    \label{fig:ciss}
\end{figure*}

\section{Related Work}
\label{sec:related}

\subsection{Unsupervised Domain Adaptation}

Previous works on UDA often utilize adversarial domain adaptation to align the source and target domains at the level of pixels, intermediate features, or outputs~\cite{FCNs:in:the:wild,SFSU_synthetic, cyCADA,adapt:structured:output:cvpr18,chen2018road,SynRealDataFogECCV18, synthetic:semantic:segmentation,FCNs:adaptation,advent:adaptation,CMAda:IJCV2020, category:adversaries:adaptation,sim:adaptation,patch:align:adaptation,decouplenet:adaptation}. Other approaches apply self-training with pseudolabels~\cite{self:training:adaptation,crst:adaptation,pixellevel:selflabeling:adaptation,simt:adaptation,diga:adaptation,IR2FRMM:adaptation} or combine self-training with adversarial adaptation~\cite{bidirectional:learning:adaptation}. CyCADA~\cite{cyCADA} employs a semantic consistency loss with some similarity to our feature invariance loss. This loss optimizes the two generators in the CycleGAN architecture~\cite{cycleGAN} to translate images across the source and target domains in a way which ensures that a \emph{fixed} segmentation network predicts the same \emph{outputs} for the translated versions of the images as for the original images. Importantly, the weights of this fixed segmentation network are not optimized jointly with the rest of the networks that are involved in CyCADA, but a separate segmentation network is rather learned for the target domain, for which no semantic consistency loss is applied. On the contrary, we propose to learn a \emph{single} segmentation network both for the source and the target domain, the \emph{internal} features of which are optimized to be invariant to the input condition. FIFO~\cite{fifo:fog:invariant:features} introduces fog factors, which are intermediate global representations of the characteristics of fog that is (or is not) present in images. These representations are extracted with a separate fog-pass filtering module, which accepts as input intermediate features of the main segmentation network. However, the fog factors\textemdash the deviation of which is penalized in~\cite{fifo:fog:invariant:features}\textemdash do not always correspond to images with the same content; thus, penalizing their deviation does not necessarily enforce condition invariance of the segmentation features.
Pixel-level adaptation via explicit transforms from source to target is performed in~\cite{fda:adaptation,textda:adaptation,dundar2018domain}; we build on the effectiveness of FDA~\cite{fda:adaptation}, but only use it as a building block in CISS, which additionally performs feature-level adaptation. Recent works upgrade the architecture and training strategy for UDA~\cite{daformer:domain:adaptation} and operate at higher resolution~\cite{hrda:domain:adaptation}, delivering significant performance gains; we implement CISS using these architectures and show the additional benefit of condition invariance in this highly competitive setting.

\subsection{Consistency Regularization}

PixMatch~\cite{pixmatch:adaptation} uses consistency regularization in the context of unsupervised domain adaptation on the target domain, by promoting invariance of the semantic predictions of the segmentation network to various perturbations of the input target image, including changes in the low-frequency part of the Fourier phase of the image and in its style. However, the original target-domain semantic predictions can be false as they constitute pseudolabels and this may impact the learned representations negatively. By contrast, our method promotes invariance of \emph{internal} features, which avoids reliance of consistency regularization on potentially false pseudolabels.
The aforementioned issue in~\cite{pixmatch:adaptation} with the reliability of pseudolabels is also present in the very recent method of MIC~\cite{mic:adaptation}, which promotes consistency in the \emph{output} space of target-domain images under masking.
The idea of consistent label predictions under input augmentations stems from FixMatch~\cite{fixmatch}, which considers a classification setting; we instead promote consistency at the level of internal features in a dense fashion. \blue{A consistency loss was also used in~\cite{bidirectional:learning:adaptation} for UDA, but it was again applied at the \emph{outputs} of the network, contrary to our feature invariance loss, which is applied to \emph{internal} network features.} Consistency under augmentations has also been found to be important in semi-supervised semantic segmentation~\cite{perturbations:semi:supervised:segmentation}; instead of plain augmentations, we employ \emph{stylization} of the input images by exploiting pairs of source and target images that are available at training to obtain better cross-domain image views for promoting invariance. CISS can be viewed as a contrastive learning method, using positive pairs to enforce feature invariance densely at each pixel. Contrastive approaches are also proposed in~\cite{memory:bank:semi:supervised:segmentation,sepico:adaptation,bidirectional:contrastive:learning:adaptation,prototypical:contrast:adaptation,CMA}, however, they contrast general pairs of pixels, while we contrast pairs of pixels that depict exactly the same point of the scene, providing stronger positive pairs. A concurrent work~\cite{siamdoge:domain:generalization} with ours implements consistency training with a consistency loss that is similar to our feature invariance loss, however, that work focuses on the domain generalization setting simply using \emph{augmentations} of input images rather than on our UDA setting, for which \emph{stylization} of the images to the style of different domains is essential. Moreover, \cite{siamdoge:domain:generalization} applies consistency to the \emph{penultimate} layer of the network, i.e.\ very close to the output level, and not \emph{internally} in the network at the encoder outputs, as CISS does. Our strategy is motivated by the intuition that the encoder of the semantic segmentation network should already output features that are invariant to the domain/style of the scene, so that the decoder can subsequently focus on parsing these features and not on further abstracting from them.

\section{Condition-Invariant Semantic Segmentation}
\label{sec:method}

We first provide a basic UDA setup for semantic segmentation, with definitions of inputs, outputs and losses, and then present our UDA method, CISS, which builds on this setup. A visual overview of CISS is presented in Fig.~\ref{fig:ciss}.

\subsection{A Basic UDA Setup}
\label{sec:method:setup}

In modern UDA training pipelines, each training batch contains an equal number $B$ of source and target images. We denote the source images by ${\{I_{s,b}\}}_{b=1}^B$ and the target images by ${\{I_{t,b}\}}_{b=1}^B$. Moreover, the batch contains pixel-level semantic labels of the source images and---in self-training-based methods---of the target images, the latter constituting pseudolabels. We denote these labels by ${\{Y_{s,b}\}}_{b=1}^B$ and ${\{\hat{Y}_{t,b}\}}_{b=1}^B$, respectively. For presenting our method, we assume that the pseudolabels ${\{\hat{Y}_{t,b}\}}_{b=1}^B$ are given, as our focus is not on improving pseudolabel generation, and we defer the details of this generation to Sec.~\ref{sec:experiments}.

For the sake of simplicity, we focus on the case where $B=1$, but our analysis extends straightforwardly to larger $B$. Dropping the redundant subscripts, the training batch in this case is $(I_s,\,I_t,\,Y_s,\,\hat{Y}_t)$. The basic UDA setup we start from involves training the semantic segmentation network $F$ using both the source-domain and the target-domain sample by applying cross-entropy losses on the outputs of $F$ for the two images. More specifically, if the semantic labels $Y$ are one-hot-encoded in a $C\times{}H\times{}W$ tensor, where $C$ is the number of classes, then the cross-entropy loss associated with the softmax output $F(I)$ of the network for $I$ is defined as
\begin{equation}
    \label{eq:cross:entropy}
    \mathcal{L}_{\text{CE}}(F, I, Y) = -\frac{1}{CHW}\sum_{c,h,w} Y_{c,h,w}\log\left({F(I)}_{c,h,w}\right).
\end{equation}
Thus, in the basic training setup we start from, the overall loss can be expressed as
\begin{equation}
    \label{eq:loss:basic}
    \mathcal{L}_{\text{basic}} = \mathcal{L}_{\text{CE}}(F, I_s, Y_s) + \mathcal{L}_{\text{CE}}(F, I_t, \hat{Y}_t).
\end{equation}
This training loss encourages the network to preserve its knowledge on semantics from the source domain, which features high-quality ground-truth labels, while also adapting to the target domain via pseudolabels.

\subsection{Pixel-Level Adaptation with Stylized Views}
\label{sec:method:stylized}

In order to better align the source and target domain, we can translate the input images from one domain to the style of the other domain. This is an alignment of the two domains at the level of pixels and it is based on the preservation of the semantic content of the input image after the stylization. Thus, the semantic annotation of the original input image can be used to supervise the prediction of the network for the stylized image, as the semantics are preserved.

This type of pixel-level adaptation has been followed in several previous works~\cite{bidirectional:learning:adaptation,synthetic:semantic:segmentation} which attempt to learn the stylization with a separate deep network. We argue that a light-weight shallow mapping $g$ for the stylization is more flexible, as stylization can be performed on-the-fly during the data loading stage of training and does not introduce unnecessary additional complexity to the overall architecture. The availability of pairs of source and target images serves such a shallow stylization well, as one image can use the other image as the reference style, so the mapping $g$ is not fixed for a given input image but has greater variability. More formally, we can write the stylized source image of our training batch from Sec.~\ref{sec:method:setup} which is computed with this regime as
\begin{equation}
    \label{eq:stylization:source}
    I_{s\to{}t} = g(I_s, I_t)
\end{equation}
and the respective stylized target image as
\begin{equation}
    \label{eq:stylization:target}
    I_{t\to{}s} = g(I_t, I_s).
\end{equation}
The stylization mapping $g$ is the same in both cases, only that the order of its arguments is flipped, as the output always has the content of the first argument and the style of the second one. Such shallow stylizations have been proposed in the color transfer work of Reinhard et al.~\cite{reinhard2001color} and in FDA~\cite{fda:adaptation} and have been shown~\cite{fda:adaptation} to perform favorably for UDA compared to stylization learned jointly with semantic segmentation. Our method is \blue{generic w.r.t.}\ the exact mapping $g$ that is used for stylization. We have used both FDA~\cite{fda:adaptation} and simple color transfer~\cite{reinhard2001color} in the implementation of CISS, motivated by the compelling results of such shallow stylization approaches, especially in the normal-to-adverse UDA setting~\cite{ACDC}. For the details of the simple color transfer method of Reinhard et al., we refer the reader to the original paper~\cite{reinhard2001color}. However, as FDA has a more complex formulation, we review it here shortly for completeness. FDA works with the discrete Fourier transform of the source and target images and copies the low-frequency Fourier amplitude of the reference style image to the input content image. More formally, FDA implements \eqref{eq:stylization:source} as
\begin{equation}
    \label{eq:stylization:source:fda}
    I_{s\to{}t} = \mathcal{F}^{-1}([M\odot{}\mathcal{F}_A(I_t)+(1-M)\odot{}\mathcal{F}_A(I_s),\mathcal{F}_P(I_s)]),
\end{equation}
where $M$ is an ideal low-pass filter, $\mathcal{F}_A(\cdot)$ denotes the Fourier amplitude, $\mathcal{F}_P(\cdot)$ denotes the Fourier phase, and $\mathcal{F}^{-1}([\cdot,\cdot])$ denotes the inverse discrete Fourier transform for a given pair of Fourier amplitude and phase. $I_{t\to{}s}$ can be computed conversely based on \eqref{eq:stylization:target}.

Since $I_{s\to{}t}$ is rendered at the style of the target domain and is thus aligned to the latter, \cite{fda:adaptation} proposes to modify the basic setup of \eqref{eq:loss:basic} and substitute the original source image $I_s$ with the stylized source image $I_{s\to{}t}$ in the cross-entropy loss associated with the source domain, where the stylization can be performed with any shallow mapping:
\begin{equation}
    \label{eq:loss:fda}
    \mathcal{L}_{\text{FDA}} = \mathcal{L}_{\text{CE}}(F, I_{s\to{}t}, Y_s) + \mathcal{L}_{\text{CE}}(F, I_t, \hat{Y}_t).
\end{equation}

\subsection{Feature Invariance Loss}
\label{sec:method:invariance}

However, by only applying cross-entropy losses on the stylized source image $I_{s\to{}t}$ and the target image $I_t$, the optimization \eqref{eq:loss:fda} proposed in~\cite{fda:adaptation} neglects the fact that \emph{two} views are available for each input image thanks to stylization, one in the style of the source domain and the other in the style of the target domain. In particular, \eqref{eq:loss:fda} only leverages the views that are characterized by the style of the target domain and neglects $I_s$ and $I_{t\to{}s}$, which are characterized by the style of the source domain. Our key insight is that by using both views of the images---each view corresponding to a different domain---in the training, we can promote \emph{invariance} across domains of the internal features generated by the network and we can thus better align the two domains at the level of features, which aids domain adaptation.

A straightforward way to attempt such an alignment is by adding cross-entropy losses on the additional views which are not included in \eqref{eq:loss:fda}, namely $I_s$ and $I_{t\to{}s}$:
\begin{align}
    \mathcal{L}_{\text{CE,full}} =\;&{\mathcal{L}}_{\text{CE}}(F, I_{s\to{}t}, Y_s) + \mathcal{L}_{\text{CE}}(F, I_s, Y_s) \nonumber\\
    &{+}\:\mathcal{L}_{\text{CE}}(F, I_t, \hat{Y}_t) + \mathcal{L}_{\text{CE}}(F, I_{t\to{}s}, \hat{Y}_t). \label{eq:loss:ce:full}
\end{align}
Since the labels used to supervise the predictions of the network for $I_s$ and $I_{s\to{}t}$ (respectively $I_t$ and $I_{t\to{}s}$) in \eqref{eq:loss:ce:full} are the same, the two predictions are indirectly attracted to the same point, which is expected to promote consistency across domains.

Nevertheless, we argue that the shared semantic content between $I_s$ and $I_{s\to{}t}$ (respectively $I_t$ and $I_{t\to{}s}$) allows to impose an even stronger constraint on the semantic segmentation network $F$. More specifically, typical deep semantic segmentation networks consist of an encoder and a decoder. The bottleneck layer between the encoder and the decoder produces high-level internal features which should ideally be invariant to the specific style or visual condition of the input, allowing the decoder to focus on parsing these features into the output semantic classes and to not have to further abstract from the specific style of the input. Thus, we can minimize the difference of internal features produced by the semantic segmentation network for views of the same scene under different styles, which is exactly the setting we have been examining. More formally, we can analyze the segmentation network $F$ as a composition of an encoder $\phi$ and a decoder $\omega$: $F=\omega\circ\phi$. For two input images $I$ and $I^{\prime}$ of the same dimensions, the features generated by the encoder are $\phi(I), \phi(I^{\prime}) \in \mathbb{R}^{D\times{}M\times{}N}$, where $D$ corresponds to the channel dimension. We define our feature invariance loss as
\begin{equation}
    \label{eq:feature:invariance}
    \mathcal{L}_{\text{inv}}(F=\omega\circ\phi, I, I^{\prime}) = \frac{1}{DMN}{\|\phi(I) - \phi(I^{\prime})\|}_\text{F}^2,
\end{equation}
where ${\|\cdot\|}_{\text{F}}$ is the Frobenius norm.

\begin{table*}[!tb]
  \caption{\textbf{Comparison of state-of-the-art domain adaptation methods on Cityscapes$\to$Dark Zurich.} The first and second groups of rows present weakly supervised methods using a RefineNet~\cite{refinenet} architecture with image-level cross-time-of-day correspondences in Dark Zurich, and unsupervised methods using a SegFormer~\cite{segformer} architecture, respectively. Best result per column in bold, second-best underlined.}
  \label{table:cs2dz}
  \centering
  \renewcommand{\arraystretch}{1.1}
  \setlength\tabcolsep{2pt}
  \footnotesize
  \smallskip
  \begin{tabular*}{\linewidth}{l @{\extracolsep{\fill}} cccccccccccccccccccc}
  \toprule
  Method & \ver{road} & \ver{sidew.} & \ver{build.} & \ver{wall} & \ver{fence} & \ver{pole} & \ver{light} & \ver{sign} & \ver{veget.} & \ver{terrain} & \ver{sky} & \ver{person} & \ver{rider} & \ver{car} & \ver{truck} & \ver{bus} & \ver{train} & \ver{motorc.} & \ver{bicycle} & mIoU\\
  \midrule
  GCMA~\cite{GCMA_UIoU:v1} & 81.7 & 46.9 & 58.8 & 22.0 & 20.0 & 41.2 & 40.5 & 41.6 & 64.8 & 31.0 & 32.1 & 53.5 & 47.5 & 75.5 & 39.2 & 0.0 & 49.6 & 30.7 & 21.0 & 42.0 \\
  MGCDA~\cite{MGCDA_UIoU} & 80.3 & 49.3 & 66.2 & 7.8 & 11.0 & 41.4 & 38.9 & 39.0 & 64.1 & 18.0 & 55.8 & 52.1 & 53.5 & 74.7 & 66.0 & 0.0 & 37.5 & 29.1 & 22.7 & 42.5 \\
  DANNet~\cite{DANNet} & 90.0 & 54.0 & 74.8 & 41.0 & \underline{21.1} & 25.0 & 26.8 & 30.2 & \underline{72.0} & 26.2 & \underline{84.0} & 47.0 & 33.9 & 68.2 & 19.0 & 0.3 & 66.4 & 38.3 & 23.6 & 44.3 \\
  \midrule
  DAFormer~\cite{daformer:domain:adaptation} & 93.5 & 65.5 & 73.3 & 39.4 & 19.2 & 53.3 & 44.1 & 44.0 & 59.5 & 34.5 & 66.6 & 53.4 & 52.7 & 82.1 & 52.7 & \phantom{0}9.5 & 89.3 & 50.5 & 38.5 & 53.8 \\
  SePiCo~\cite{sepico:adaptation} & 93.2 & 68.1 & 73.7 & 32.8 & 16.3 & 54.6 & \underline{49.5} & 48.1 & \textbf{74.2} & 31.0 & \textbf{86.3} & 57.9 & 50.9 & 82.4 & 52.2 & \phantom{0}1.3 & 83.8 & 43.9 & 29.8 & 54.2 \\
  HRDA~\cite{hrda:domain:adaptation} & 90.4 & 56.3 & 72.0 & 39.5 & 19.5 & 57.8 & \textbf{52.7} & 43.1 & 59.3 & 29.1 & 70.5 & 60.0 & \underline{58.6} & \underline{84.0} & \underline{75.5} & 11.2 & 90.5 & 51.6 & 40.9 & 55.9 \\
  MIC~\cite{mic:adaptation} & \textbf{94.8} & \textbf{75.0} & \textbf{84.0} & \textbf{55.1} & \textbf{28.4} & \textbf{62.0} & 35.5 & \underline{52.6} & 59.2 & \textbf{46.8} & 70.0 & \textbf{65.2} & \textbf{61.7} & 82.1 & 64.2 & \textbf{18.5} & \underline{91.3} & \underline{52.6} & \underline{44.0} & \underline{60.2} \\
  CISS (ours) & \underline{94.3} & \underline{70.4} & \underline{80.7} & \underline{50.8} & 20.9 & \underline{59.1} & 36.1 & \textbf{57.3} & 67.9 & \underline{37.5} & 82.7 & \underline{62.9} & 55.7 & \textbf{85.7} & \textbf{83.5} & \underline{14.0} & \textbf{91.8} & \textbf{55.4} & \textbf{45.9} & \textbf{60.7} \\
  \bottomrule
  \end{tabular*}
\end{table*}

Coming back to our UDA setup, we propose to apply our feature invariance loss on the pairs of views $(I_s, I_{s\to{}t})$ and $(I_t, I_{t\to{}s})$ in order to align the internal features of the views from each pair. The two resulting feature invariance losses are combined with the cross-entropy losses of the basic setup of \eqref{eq:loss:basic} in our final formulation of CISS as
\begin{align}
    \mathcal{L}_{\text{CISS}} =\;&{\mathcal{L}}_{\text{CE}}(F, I_s, Y_s) + \mathcal{L}_{\text{CE}}(F, I_t, \hat{Y}_t) \nonumber\\
    &{+}\:\lambda_s\mathcal{L}_{\text{inv}}(F, I_s, I_{s\to{}t}) + \lambda_t\mathcal{L}_{\text{inv}}(F, I_t, I_{t\to{}s}), \label{eq:loss:ciss}
\end{align}
where $\lambda_s$ and $\lambda_t$ are tunable hyperparameters. Note that we use cross-entropy losses only on the original images $I_s$ and $I_t$, as the cross-entropy losses on the stylized images $I_{s\to{}t}$ and $I_{t\to{}s}$ which are used in \eqref{eq:loss:ce:full} are redundant due to the inclusion of the feature invariance losses. In Sec.~\ref{sec:experiments}, we thoroughly ablate the final formulation in \eqref{eq:loss:ciss} and compare it to the basic formulation in \eqref{eq:loss:basic} and the alternative formulations in \eqref{eq:loss:fda} and \eqref{eq:loss:ce:full}, demonstrating the benefit of introducing our novel feature invariance loss compared to training with the other formulations.

\section{Experiments}
\label{sec:experiments}

\subsection{Experimental Setup}
\label{sec:experiments:setup}

\subsubsection{Implementation Details}

The default implementation of CISS is based on HRDA~\cite{hrda:domain:adaptation}. Our semantic segmentation network comprises an MiT-B5 encoder from SegFormer~\cite{segformer} and a context-aware feature fusion decoder~\cite{daformer:domain:adaptation}. We also implement CISS with a DeepLabv2~\cite{DeepLab:v2} architecture involving a ResNet-101 backbone~\cite{resnet}, in order to compare directly to several earlier UDA methods which use this architecture. For the default HRDA-based implementation, we follow the teacher-student self-training framework of DAFormer~\cite{daformer:domain:adaptation} with confidence-weighted pseudolabels, rare class sampling, and target data augmentation following DACS~\cite{dacs:domain:adaptation}, and we use the AdamW optimizer~\cite{adamw} with a learning rate of $6\times{}10^{-5}$ for the encoder and $6\times{}10^{-4}$ for the decoder, a linear learning rate warm-up, and mini-batches of size $B=2$. We follow the default configuration and parameters of HRDA regarding its multi-resolution setup. Unless otherwise stated, we use FDA~\cite{fda:adaptation} by default \blue{as the stylization module $g$.} Alternatively and only when explicitly stated, we \blue{instantiate $g$ with} the color transfer of \cite{reinhard2001color} \blue{or with simple color jitter augmentation. In the latter, we randomly perturb independently with 50\% probability each of the brightness, contrast, saturation, and hue of the input image. In all cases, our stylization operation $g$ is always applied both to source-domain and target-domain data, while DACS-based augmentation is applied only to target-domain data, only with a probability---i.e.\ not always---and after stylization $g$. Thus, the encoder $\phi$ is encouraged by CISS to become invariant w.r.t.\ style variations under mapping $g$ \emph{per se}. DACS-based augmentation does not interfere with the former learning objective but rather synergizes with it by orthogonally improving target-domain pseudolabels.} In the application of FDA stylization, we use $\beta=0.01$ as the bandwidth parameter of the low-frequency band of the Fourier spectrum, \blue{following the default choice in the original paper~\cite{fda:adaptation}.} We set the default values of the weights of the feature invariance losses in \eqref{eq:loss:ciss} for adaptation from Cityscapes to ACDC to $\lambda_s = 50$ and $\lambda_t = 20$ for the default HRDA-based implementation of CISS and to $\lambda_s = \lambda_t = 10$ for the alternative DeepLabv2-based implementation. For adaptation from Cityscapes to Dark Zurich, we set $\lambda_s = 100$ and $\lambda_t = 10$. We provide a study of these weights in Sec.~\ref{sec:experiments:ablations}.

\begin{table*}[!tb]
  \caption{\textbf{Comparison of state-of-the-art unsupervised domain adaptation methods on Cityscapes$\to$ACDC.} Cityscapes serves as the source domain and the entire ACDC including all four adverse conditions serves as the target domain. The first, second and third groups of rows present methods trained externally on Cityscapes$\to$Dark Zurich, DeepLabv2-based UDA methods and SegFormer-based UDA methods, respectively. Results of DACS are taken from~\cite{refign:domain:adaptation}. Best result per column in bold, second-best underlined.}
  \label{table:cs2acdc:sota:all}
  \centering
  \renewcommand{\arraystretch}{1.1}
  \setlength\tabcolsep{2pt}
  \footnotesize
  \smallskip
  \begin{tabular*}{\linewidth}{l @{\extracolsep{\fill}} cccccccccccccccccccc}
  \toprule
  Method & \ver{road} & \ver{sidew.} & \ver{build.} & \ver{wall} & \ver{fence} & \ver{pole} & \ver{light} & \ver{sign} & \ver{veget.} & \ver{terrain} & \ver{sky} & \ver{person} & \ver{rider} & \ver{car} & \ver{truck} & \ver{bus} & \ver{train} & \ver{motorc.} & \ver{bicycle} & mIoU\\
  \midrule
  GCMA~\cite{GCMA_UIoU:v1} & 79.7 & 48.7 & 71.5 & 21.6 & 29.9 & 42.5 & 56.7 & 57.7 & \underline{75.8} & 39.5 & \underline{87.2} & 57.4 & 29.7 & 80.6 & 44.9 & 46.2 & 62.0 & 37.2 & 46.5 & 53.4\\
  MGCDA~\cite{MGCDA_UIoU} & 76.0 & 49.4 & 72.0 & 11.3 & 21.7 & 39.5 & 52.0 & 54.9 & 73.7 & 24.7 & \textbf{88.6} & 54.1 & 27.2 & 78.2 & 30.9 & 41.9 & 58.2 & 31.1 & 44.4 & 48.9\\
  \midrule
  AdaptSegNet~\cite{adapt:structured:output:cvpr18} & 69.4 & 34.0 & 52.8 & 13.5 & 18.0 & \phantom{0}4.3 & 14.9 & \phantom{0}9.7 & 64.0 & 23.1 & 38.2 & 38.6 & 20.1 & 59.3 & 35.6 & 30.6 & 53.9 & 19.8 & 33.9 &    33.4\\
  BDL~\cite{bidirectional:learning:adaptation} & 56.0 & 32.5 & 68.1 & 20.1 & 17.4 & 15.8 & 30.2 & 28.7 & 59.9 & 25.3 & 37.7 & 28.7 & 25.5 & 70.2 & 39.6 & 40.5 & 52.7 & 29.2 & 38.4 &      37.7\\
  CLAN~\cite{category:adversaries:adaptation} & 79.1 & 29.5 & 45.9 & 18.1 & 21.3 & 22.1 & 35.3 & 40.7 & 67.4 & 29.4 & 32.8 & 42.7 & 18.5 & 73.6 & 42.0 & 31.6 & 55.7 & 25.4 & 30.7 &      39.0\\
  CRST~\cite{crst:adaptation} & 51.7 & 24.4 & 67.8 & 13.3 & \phantom{0}9.7 & 30.2 & 38.2 & 34.1 & 58.0 & 25.2 & 76.8 & 39.9 & 17.1 & 65.4 & \phantom{0}3.7 & \phantom{0}6.6 & 39.6 & 11.8 & 8.6 &     32.8\\
  FDA~\cite{fda:adaptation} & 73.2 & 34.7 & 59.0 & 24.8 & 29.5 & 28.6 & 43.3 & 44.9 & 70.1 & 28.2 & 54.7 & 47.0 & 28.5 & 74.6 & 44.8 & 52.3 & 63.3 & 28.3 & 39.5 &    45.7\\
  SIM~\cite{sim:adaptation} & 53.8 & 6.8 & 75.5 & 11.6 & 22.3 & 11.7 & 23.4 & 25.7 & 66.1 & \phantom{0}8.3 & 80.6 & 41.8 & 24.8 & 49.7 & 38.6 & 21.0 & 41.8 & 25.1 & 29.6 &    34.6\\
  MRNet~\cite{mrnet:rectifying} & 72.2 & \phantom{0}8.2 & 36.4 & 13.7 & 18.5 & 20.4 & 38.7 & 45.4 & 70.2 & 35.7 & \phantom{0}5.0 & 47.8 & 19.1 & 73.6 & 42.1 & 36.0 & 47.4 & 17.7 & 37.4 &    36.1\\
  DACS~\cite{dacs:domain:adaptation} & 58.5 & 34.7 & 76.4 & 20.9 & 22.6 & 31.7 & 32.7 & 46.8 & 58.7 & 39.0 & 36.3 & 43.7 & 20.5 & 72.3 & 39.6 & 34.8 & 51.1 & 24.6 & 38.2 & 41.2\\
  CISS-DeepLabv2 (ours) & 70.5 & 36.7 & 67.0 & 29.4 & 30.2 & 31.6 & 45.6 & 48.9 & 70.4 & 24.7 & 65.5 & 48.2 & 31.1 & 76.6 & 45.7 & 47.0 & 62.8 & 26.8 & 38.9 & 47.2\\
  \midrule
  DAFormer~\cite{daformer:domain:adaptation} & 58.4 & 51.3 & 84.0 & 42.7 & 35.1 & 50.7 & 30.0 & 57.0 & 74.8 & 52.8 & 51.3 & 58.2 & 32.6 & 82.7 & 58.3 & 54.9 & 82.4 & 44.1 & 50.7 & 55.4\\
  SePiCo~\cite{sepico:adaptation} & 61.3 & 48.6 & 84.9 & 39.6 & 40.3 & 54.2 & 48.9 & 60.6 & 74.8 & 54.3 & 57.2 & 65.2 & 38.3 & 84.8 & 66.2 & 60.4 & 85.5 & 44.5 & 53.1 & 59.1\\
  HRDA~\cite{hrda:domain:adaptation} & 88.3 & 57.9 & 88.1 & \underline{55.2} & 36.7 & \underline{56.3} & \underline{62.9} & 65.3 & 74.2 & 57.7 & 85.9 & 68.8 & 45.6 & 88.5 & \textbf{76.4} & \underline{82.4} & \underline{87.7} & 52.7 & 60.4 & 68.0\\
  MIC~\cite{mic:adaptation} & \underline{90.8} & \underline{67.1} & \textbf{89.2} & 54.5 & \textbf{40.5} & \textbf{57.2} & 62.0 & \textbf{68.4} & \textbf{76.3} & \textbf{61.8} & 87.0 & \textbf{71.3} & \textbf{49.4} & \textbf{89.7} & \underline{75.7} & \textbf{86.8} & \textbf{89.1} & \textbf{56.9} & \textbf{63.0} & \textbf{70.4} \\
  CISS (ours) & \textbf{92.0} & \textbf{69.6} & \textbf{89.2} & \textbf{57.3} & \textbf{40.5} & 55.8 & \textbf{67.1} & \underline{67.3} & 75.3 & \underline{59.7} & 86.4 & \underline{70.0} & \underline{47.5} & \underline{88.9} & 73.1 & 77.5 & 87.0 & \underline{55.6} & \underline{61.7} & \underline{69.6}\\
  \bottomrule
  \end{tabular*}
\end{table*}

\begin{figure*}
  \centering
  \vspace{-0.35cm}
  \subfloat{\includegraphics[width=0.248\textwidth]{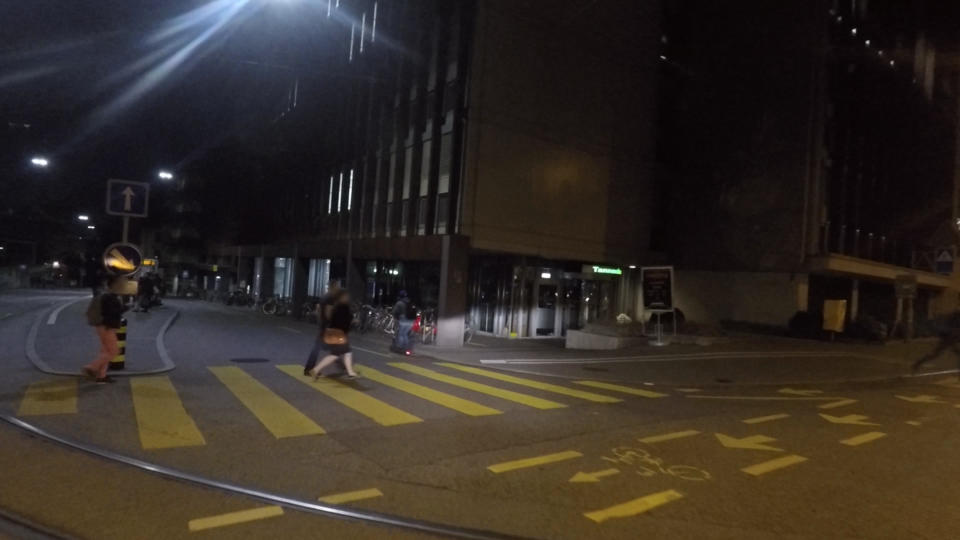}}
  \hfil
  \subfloat{\includegraphics[width=0.248\textwidth]{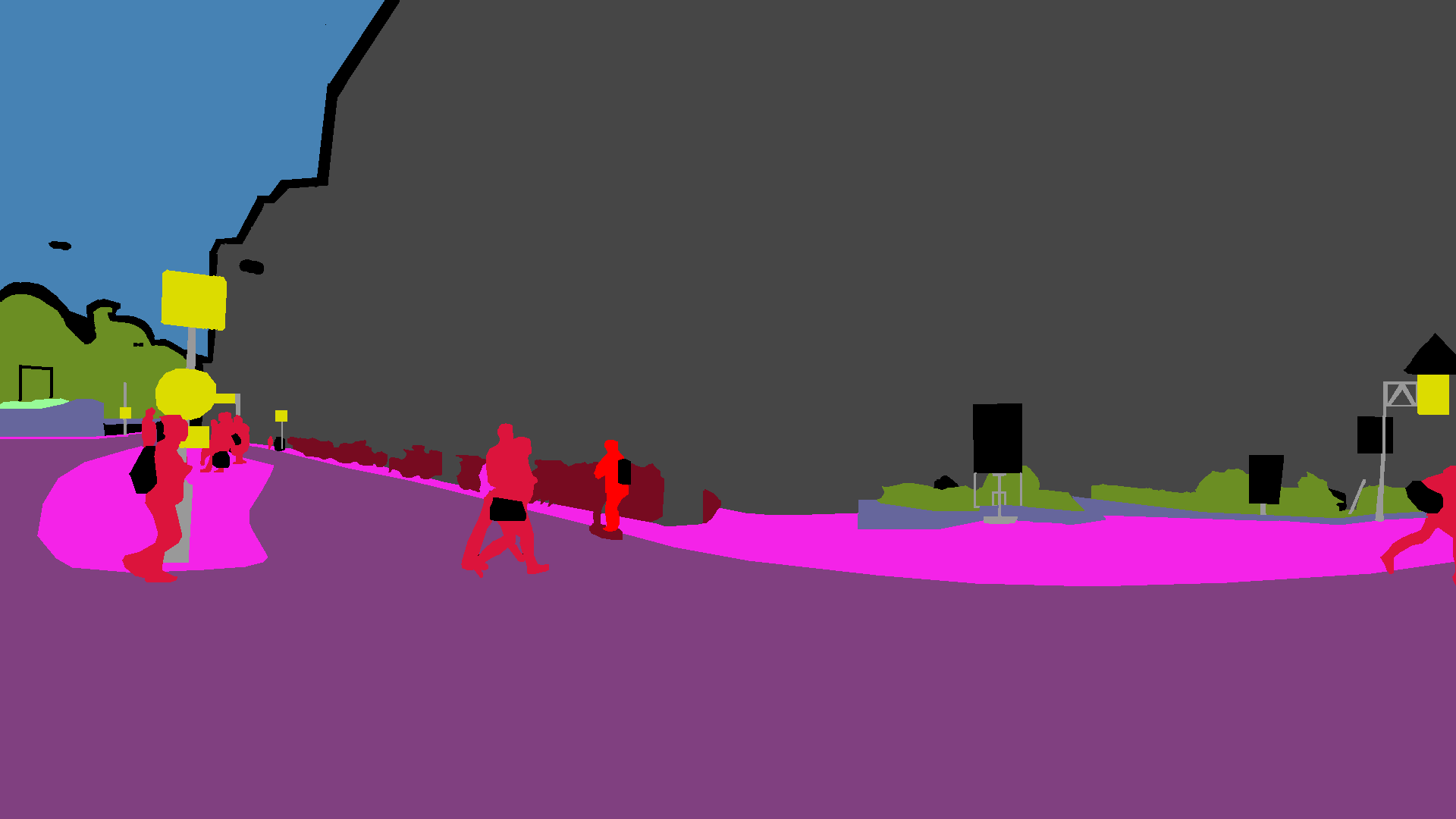}}
  \hfil
  \subfloat{\includegraphics[width=0.248\textwidth]{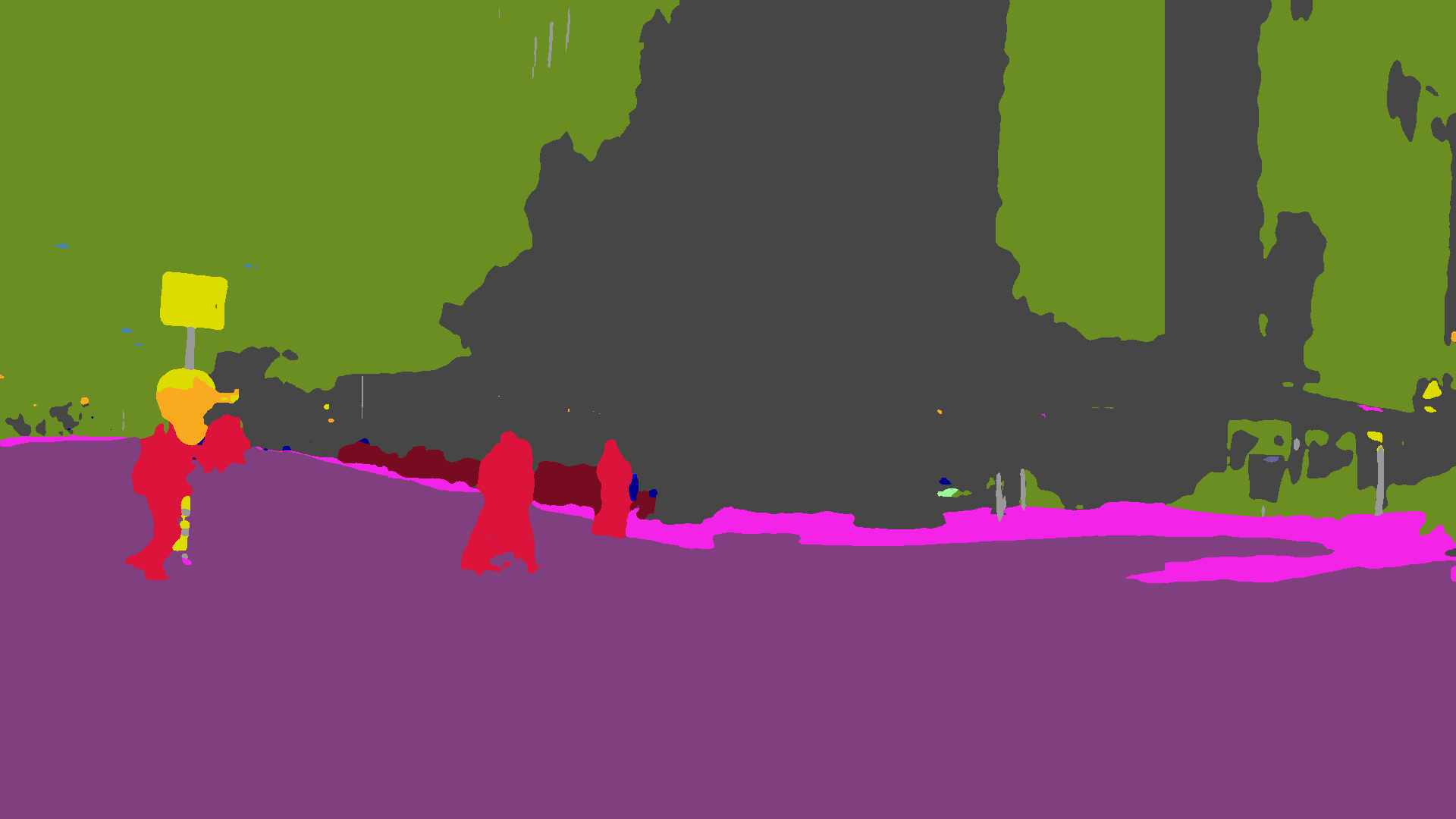}}
  \hfil
  \subfloat{\includegraphics[width=0.248\textwidth]{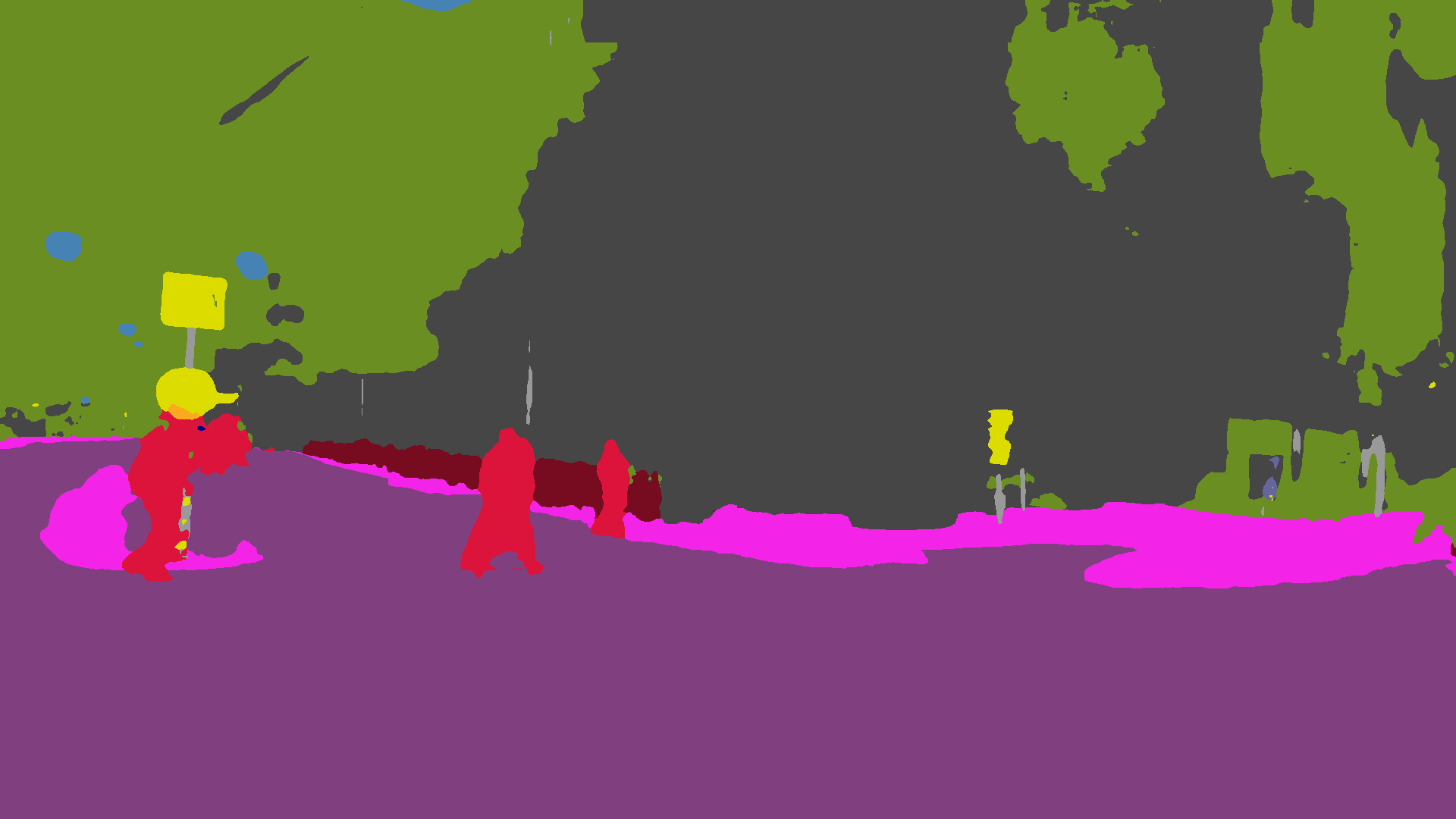}}
  \\
  \vspace{-0.35cm}
  \subfloat{\includegraphics[width=0.248\textwidth]{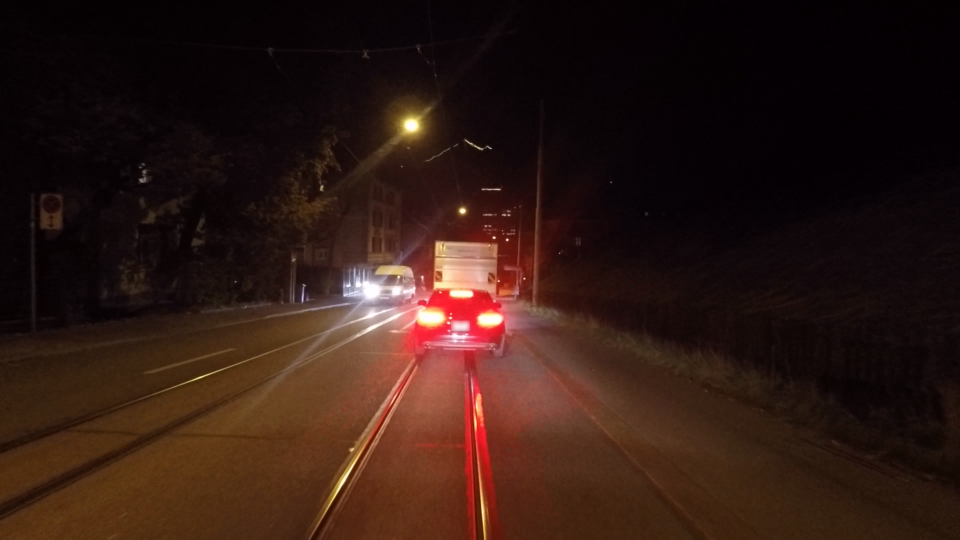}}
  \hfil
  \subfloat{\includegraphics[width=0.248\textwidth]{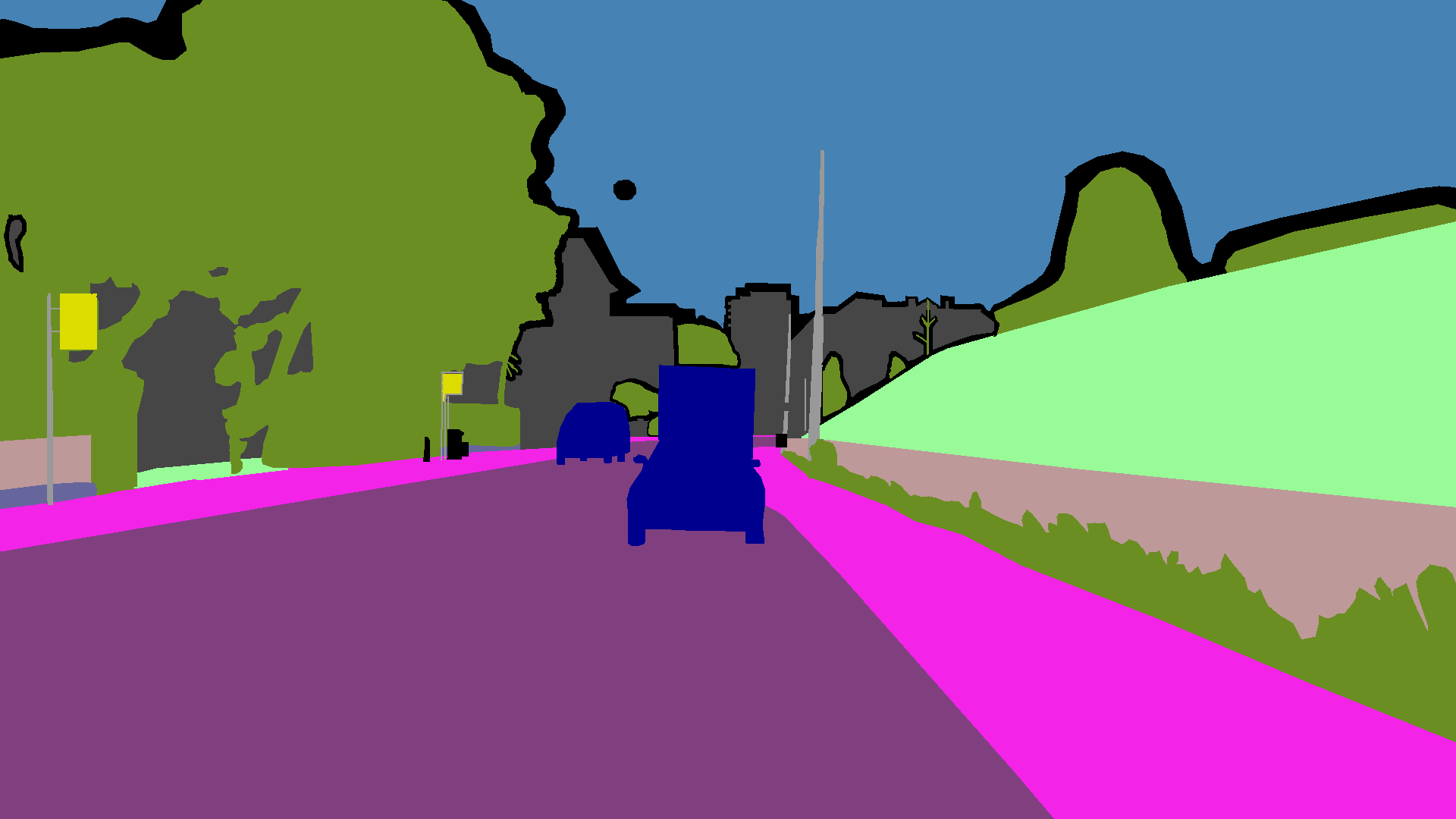}}
  \hfil
  \subfloat{\includegraphics[width=0.248\textwidth]{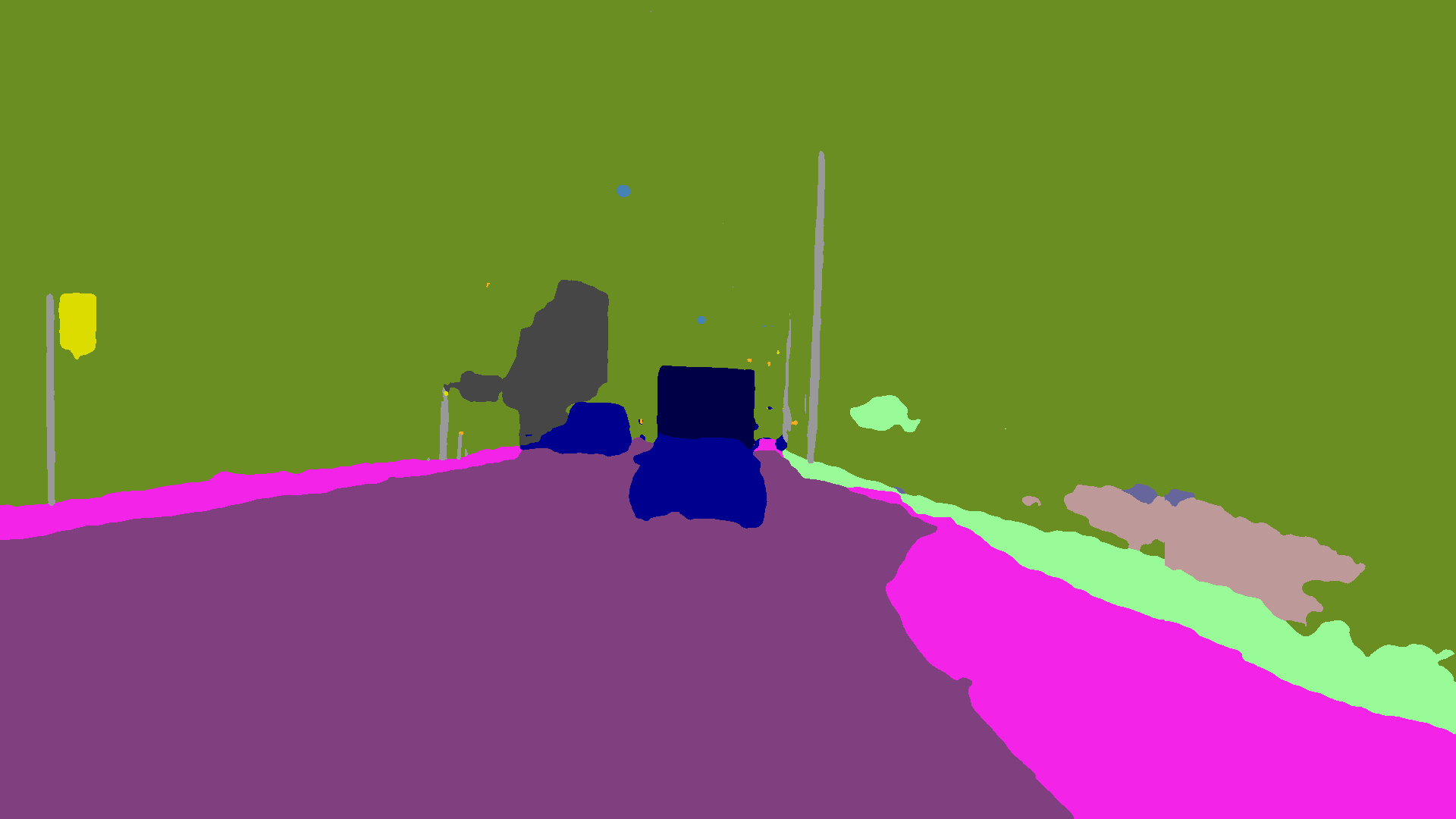}}
  \hfil
  \subfloat{\includegraphics[width=0.248\textwidth]{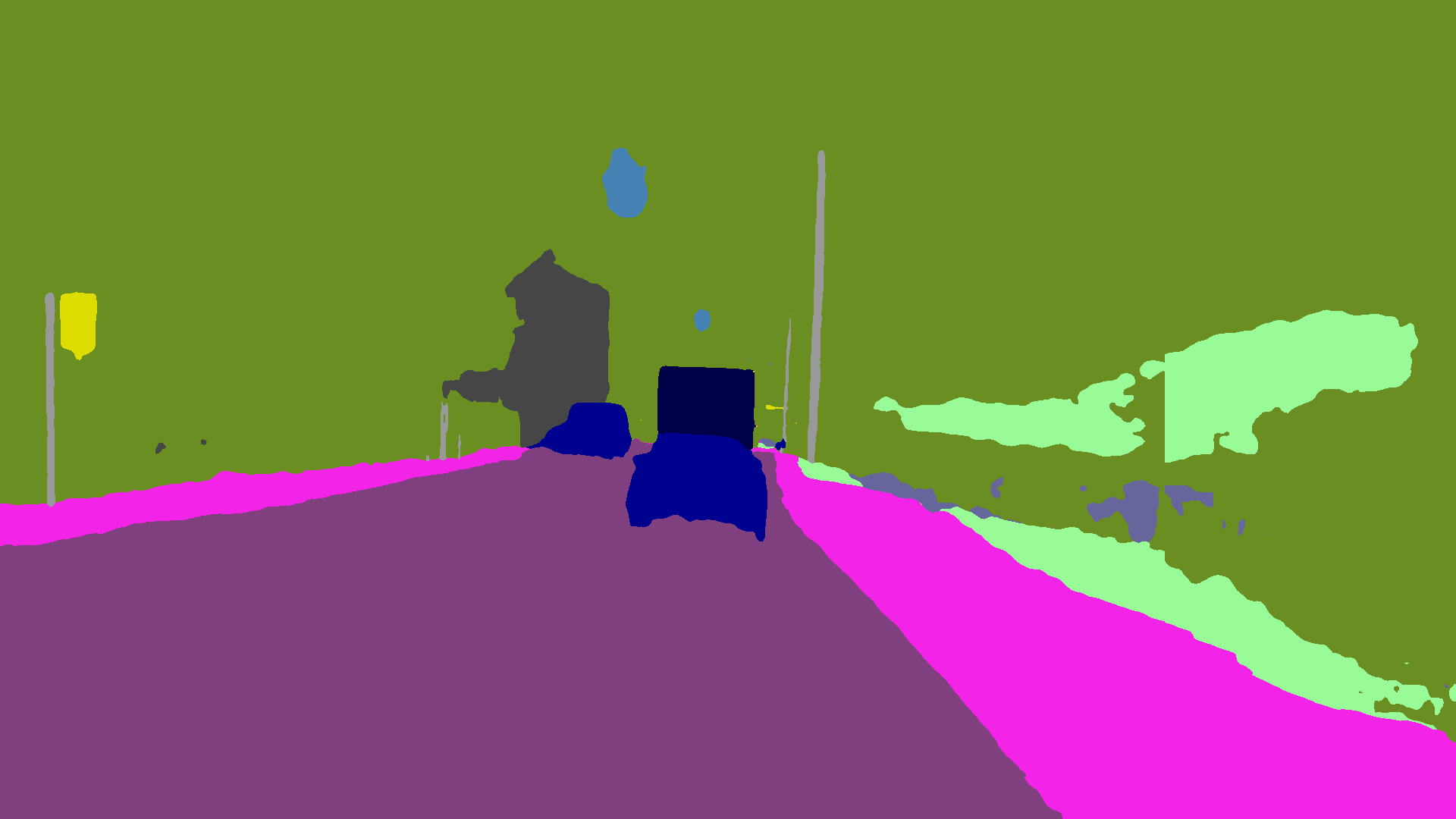}}
  \\
  \vspace{-0.35cm}
  \subfloat{\includegraphics[width=0.248\textwidth]{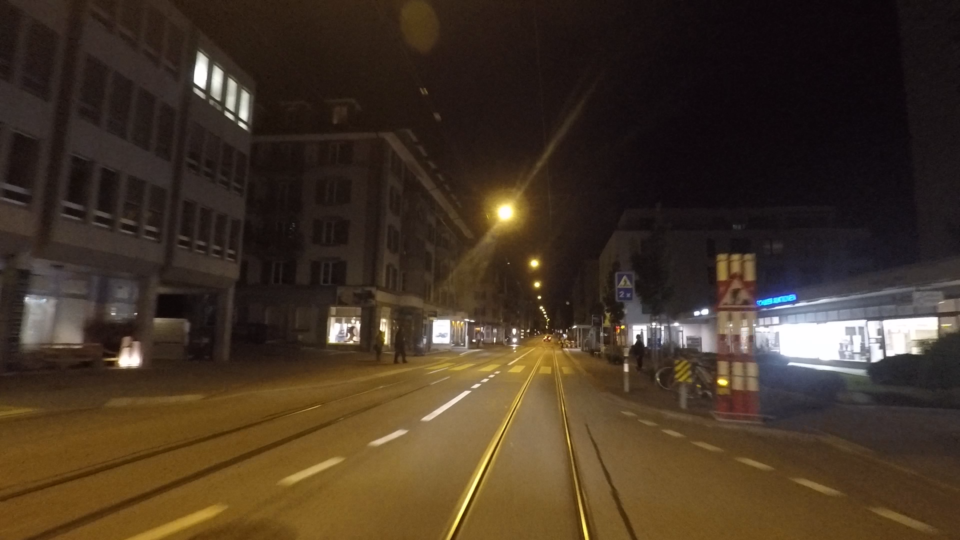}}
  \hfil
  \subfloat{\includegraphics[width=0.248\textwidth]{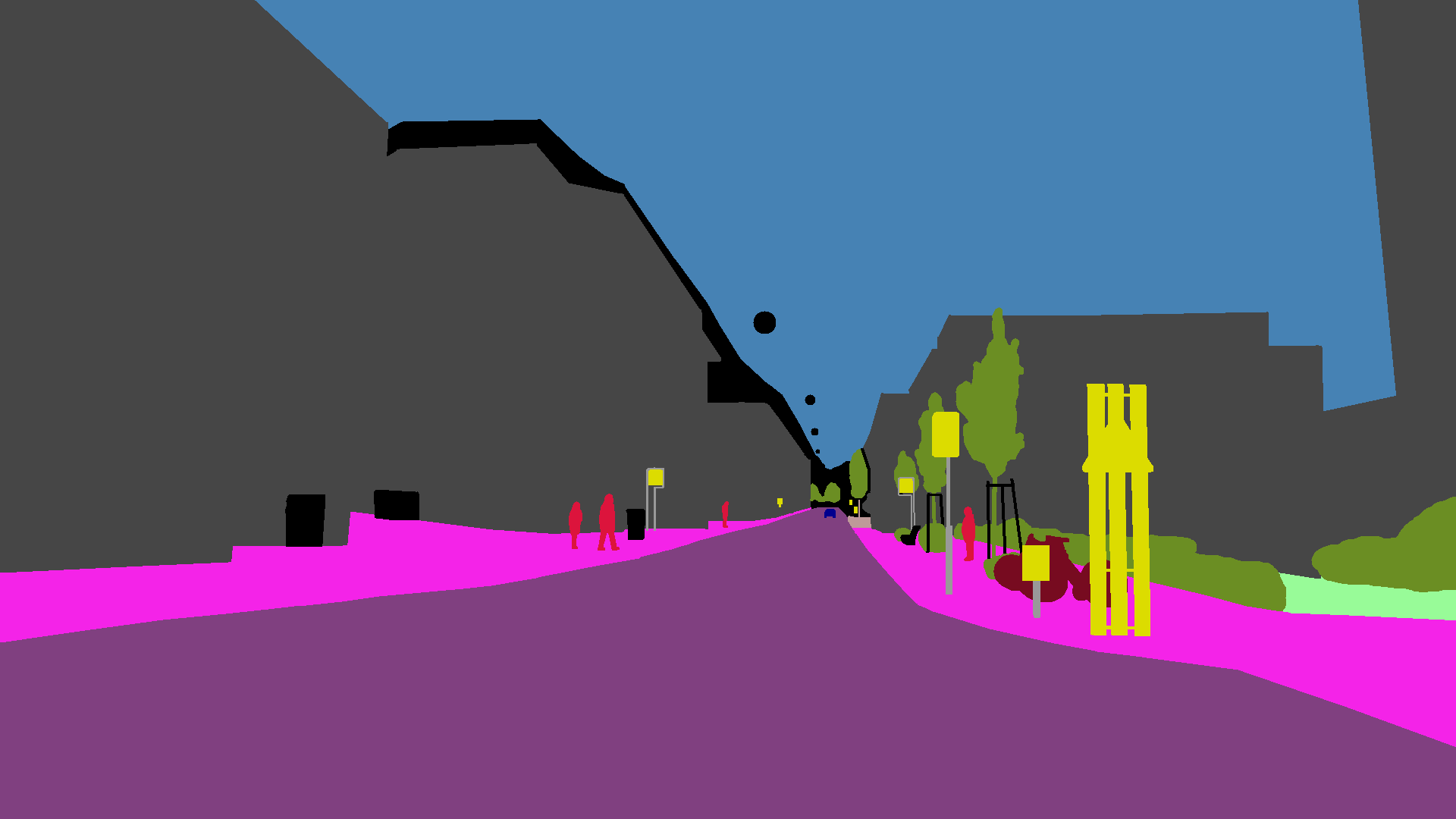}}
  \hfil
  \subfloat{\includegraphics[width=0.248\textwidth]{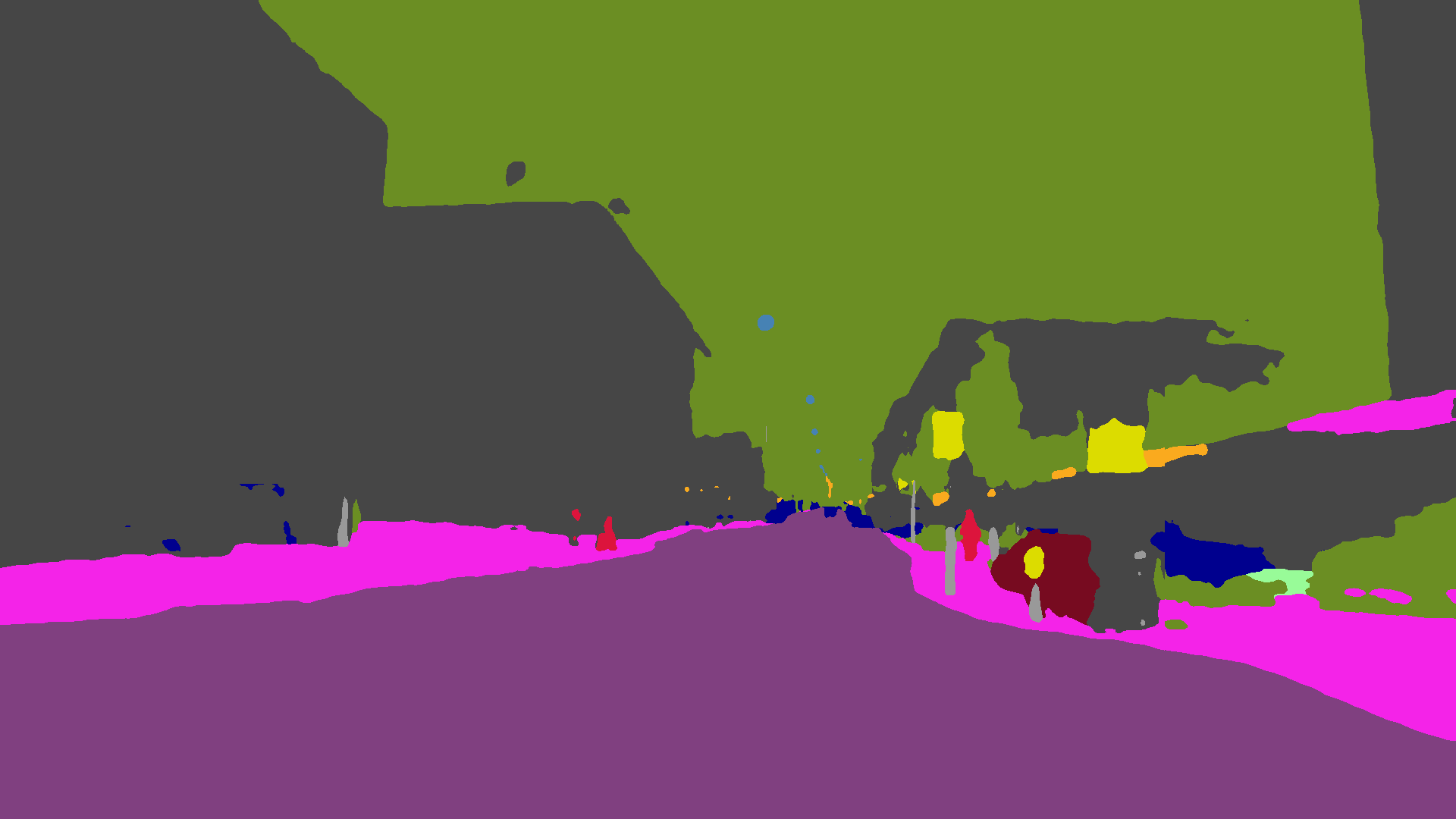}}
  \hfil
  \subfloat{\includegraphics[width=0.248\textwidth]{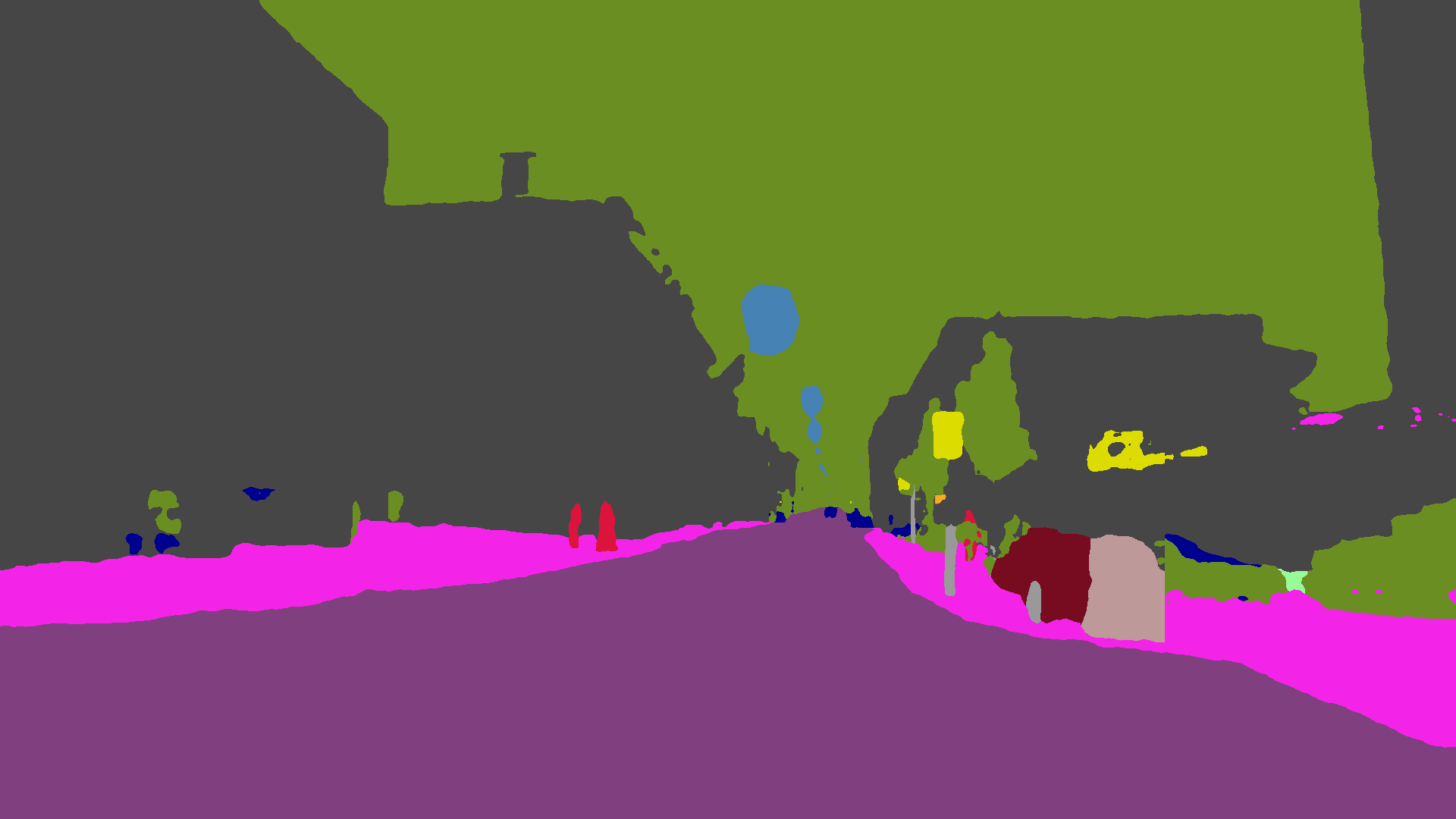}}
  \\
  \vspace{-0.35cm}
  \subfloat{\includegraphics[width=0.248\textwidth]{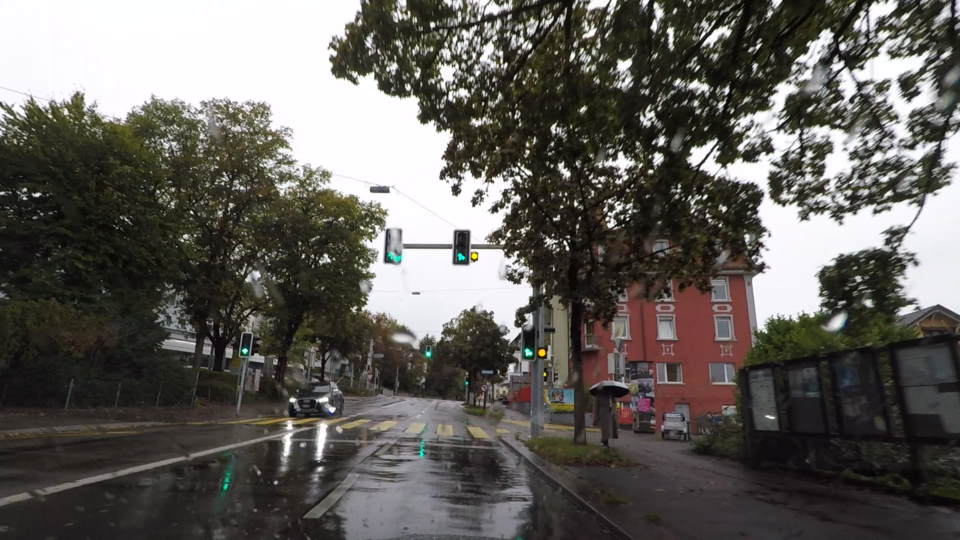}}
  \hfil
  \subfloat{\includegraphics[width=0.248\textwidth]{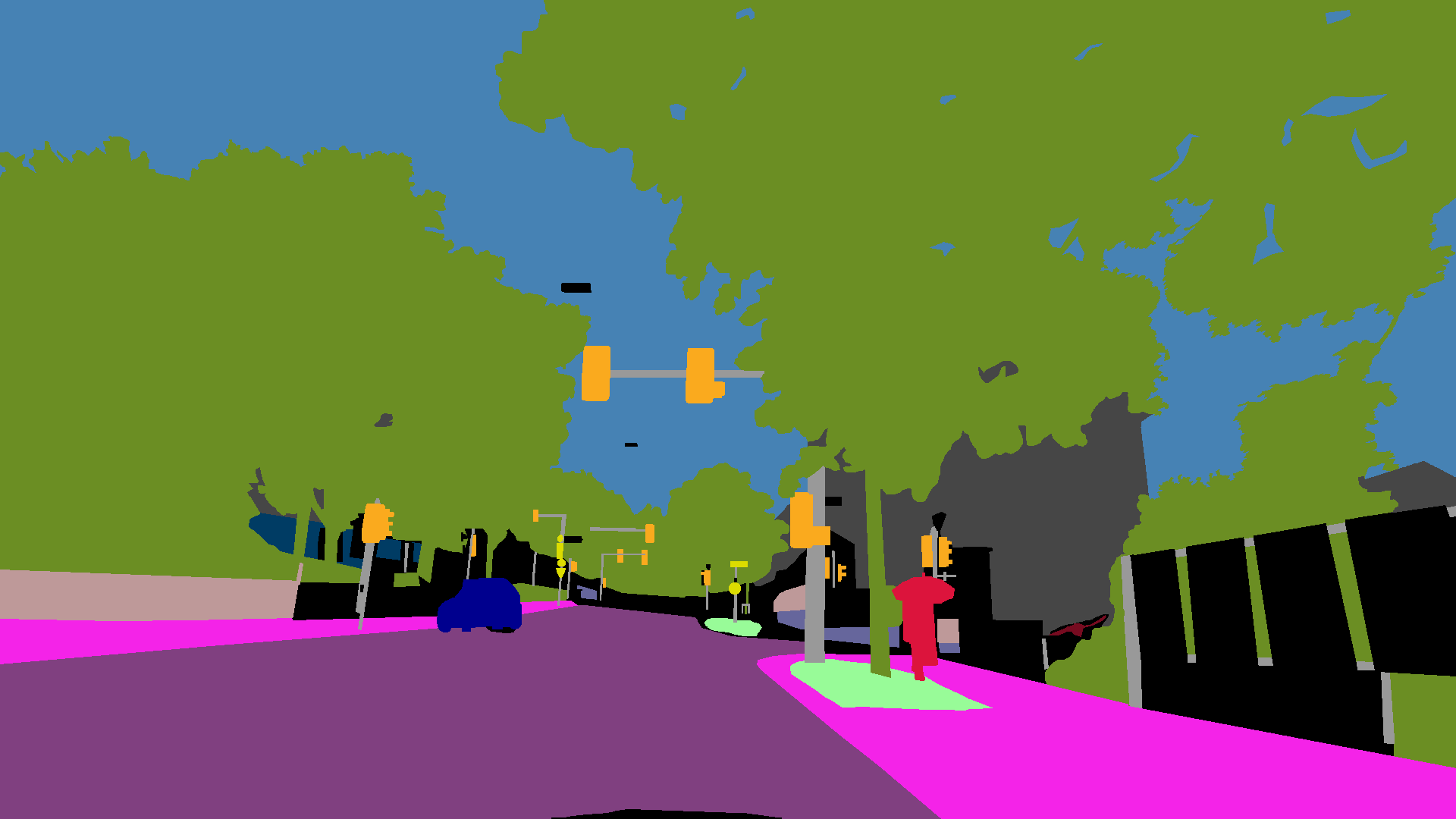}}
  \hfil
  \subfloat{\includegraphics[width=0.248\textwidth]{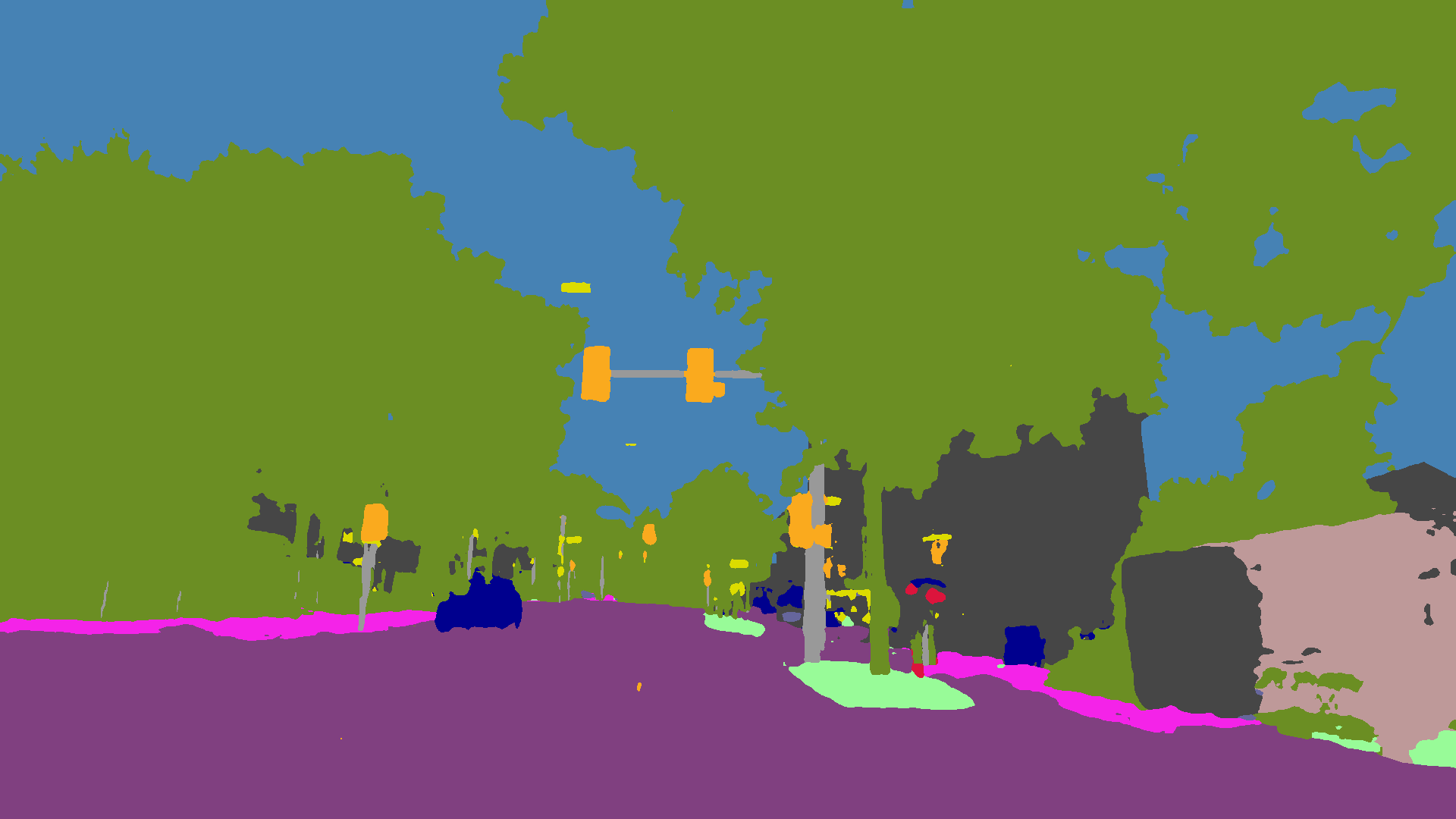}}
  \hfil
  \subfloat{\includegraphics[width=0.248\textwidth]{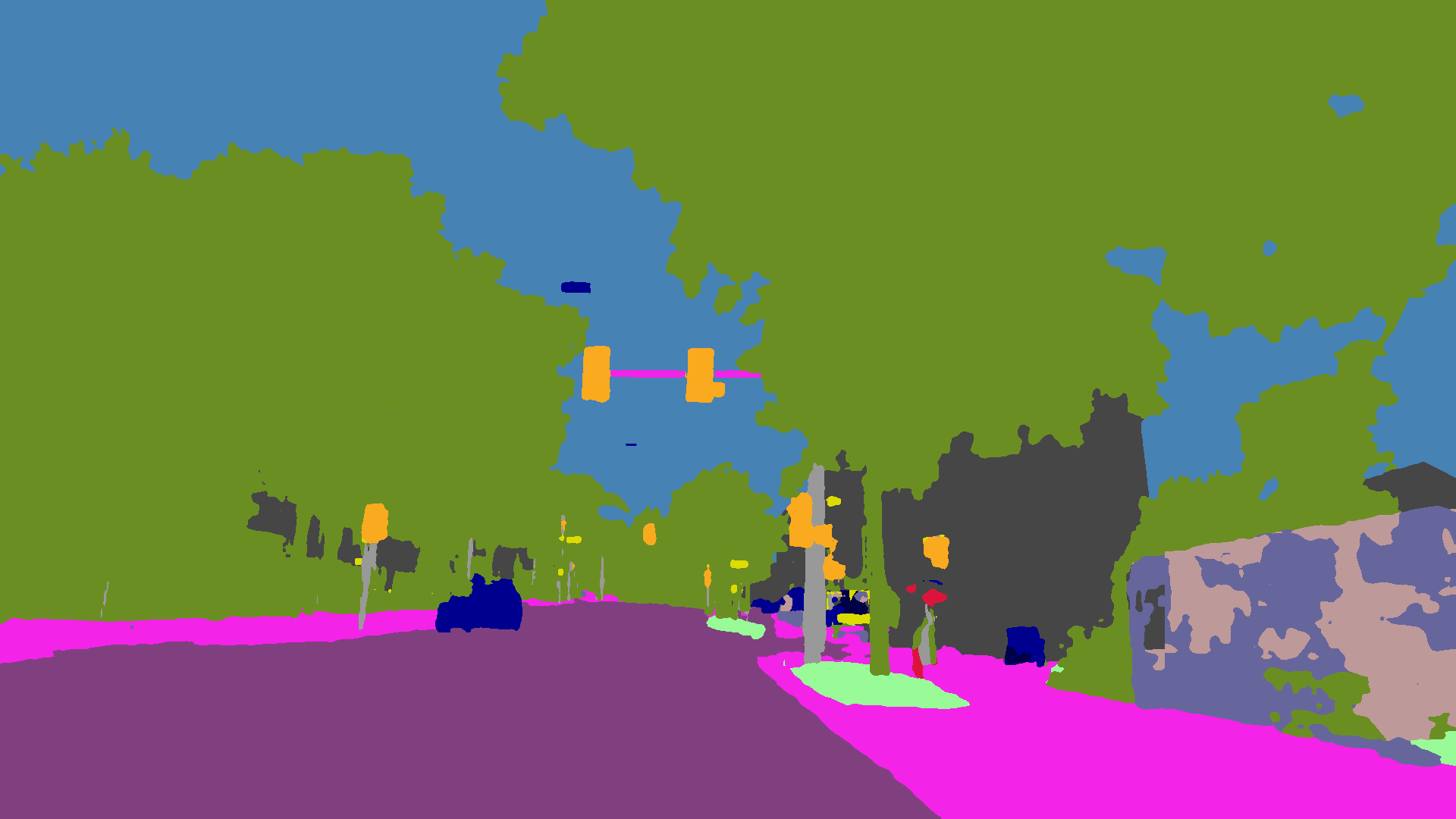}}
  \caption{\textbf{Qualitative results on Cityscapes$\to{}$ACDC.} From left to right: ACDC image, ground-truth annotation, HRDA~\cite{hrda:domain:adaptation}, and CISS. Best viewed on a screen and zoomed in.}
  \label{fig:comparison:acdc}
\end{figure*}

\subsubsection{Datasets}

In our experiments, we focus on the setting of domain adaptation and generalization from normal to adverse visual conditions, as our method is tailored for condition-level domain shifts that affect the appearance and texture of objects in the scene and not for structural-level shifts, as in the synthetic-to-real scenario. We use Cityscapes~\cite{Cityscapes} as the labeled source-domain set in our experiments. Cityscapes is a large dataset of urban driving scenes, captured in several cities of central Europe under normal conditions and containing high-quality pixel-level semantic annotations for a set of 19 common classes in driving scenes. It consists of a training set with 2975 images, a validation set with 500 images, and a test set with 1525 images. When training UDA methods in our experiments, we sample source images only from the training set of Cityscapes. In addition, we use Dark Zurich~\cite{MGCDA_UIoU} and ACDC~\cite{ACDC} as unlabeled target-domain sets, which model the adverse-condition domain for normal-to-adverse UDA. Dark Zurich comprises 2617 nighttime images of driving scenes, which are split into 2416 training, 50 validation, and 151 test images. ACDC consists of 4006 images of driving scenes distributed evenly among four common adverse conditions, i.e., night, fog, rain, and snow. Its training, validation and test set contain 1600, 406 and 2000 images respectively. Both Dark Zurich and ACDC feature high-quality semantic annotations for the same set of 19 classes as Cityscapes. In our experiments, we use the training sets of Dark Zurich and ACDC as the unlabeled target training sets, evaluate on the respective validation sets for ablations and hyperparameter studies, and evaluate only our final models against competing methods on the respective test sets, both of which have withheld annotations and thus serve as competitive public UDA benchmarks. Finally, for models adapted to nighttime segmentation on Dark Zurich, we use BDD100K-night~\cite{MGCDA_UIoU,BDD100K} \blue{and ACDC-night~\cite{ACDC}} as target \blue{sets} for generalization. BDD100K-night consists of 87 nighttime images with accurate segmentation labels and is a subset of the BDD100K segmentation dataset~\cite{BDD100K}. \blue{ACDC-night is the nighttime part of the test set of ACDC with 500 challenging nighttime images and has an associated specialized public nighttime benchmark~\cite{ACDC} based on its withheld ground-truth labels.}

\subsection{Comparison to the State of the Art in UDA}
\label{sec:experiments:sota}

A general note regarding certain state-of-the-art UDA methods that we compare against is that MIC~\cite{mic:adaptation} is also built on top of HRDA~\cite{hrda:domain:adaptation}, as is the case with CISS. However, CISS and MIC improve upon their common HRDA baseline along orthogonal methodological directions. These facts imply that when the performance of CISS is comparable with that of MIC, each of the two methods has independently improved by a similar margin over HRDA. Moreover, even slight performance gains of CISS over MIC are significant, as they are achieved on top of the existing improvement of MIC over HRDA, with the latter being the starting point of CISS too.

\subsubsection{Cityscapes$\to$Dark Zurich}
\label{sec:experiments:sota:cs2dz}

We present the comparison of CISS to competing state-of-the-art domain adaptation methods on the challenging daytime-to-nighttime Cityscapes$\to$Dark Zurich domain adaptation benchmark in Table~\ref{table:cs2dz}. In particular, we compare CISS both to more directly related SegFormer-based UDA methods and to weakly supervised domain adaptation methods which additionally utilize during training the cross-time-of-day correspondences which are available in Dark Zurich. CISS outperforms all other methods and sets the new state of the art for UDA on Cityscapes$\to$Dark Zurich, with a mean IoU of 60.7\%. More specifically, CISS improves by a substantial 4.8\% over its baseline, HRDA. This improvement is greater than the respective improvement of the previous state-of-the-art method, MIC, over HRDA, rendering CISS better than MIC overall. On top of that, CISS exhibits a more stable performance across different classes than MIC, as CISS scores more than 10\% lower in class-level IoU than the respective top method for only one class, whereas the same deficit occurs for four classes for MIC. A further comparison of the models which are evaluated in Table~\ref{table:cs2dz} in a zero-shot generalization setting is presented in Sec.~\ref{sec:experiments:generalization}.

\subsubsection{Cityscapes$\to$ACDC}
\label{sec:experiments:sota:cs2acdc}

We present the comparison of CISS to competing state-of-the-art UDA methods on Cityscapes$\to$ACDC adaptation in Table~\ref{table:cs2acdc:sota:all}. CISS and MIC are the two best methods and consistently outperform other methods in most classes, with MIC scoring slightly higher in mean IoU than CISS. Moreover, both CISS and MIC consistently outperform their common baseline, HRDA, in most classes: CISS beats HRDA in 15/19 classes, while MIC beats HRDA in 16/19 classes.
Our method achieves the best or second-best IoU in 13/19 individual classes, excelling in classes that are crucial for driving perception, such as road, sidewalk, traffic light, person, and car.
In terms of pixel accuracy, which is another widely used metric in semantic segmentation beyond mean IoU, CISS is the top-performing method along with MIC on Cityscapes$\to$ACDC, as the two methods are on a par at 90.3\%. Thus, CISS, which represents an independent and orthogonal domain adaptation strategy to MIC, improves upon the common HRDA baseline---which has a pixel accuracy of 89.1\%---equally significantly to MIC, in terms of both mean IoU and pixel accuracy.
Focusing on the methods that use a DeepLabv2 architecture, CISS-DeepLabv2 has the top performance among them, showing that the benefit of our novel feature invariance loss is general across different UDA architectures.

Qualitative comparisons of CISS to its baseline, i.e.\ HRDA~\cite{hrda:domain:adaptation}, on Cityscapes$\to$ACDC are presented in Fig.~\ref{fig:comparison:acdc}, showing segmentation results on validation images of ACDC. On the top nighttime image, our method successfully segments both the traffic signs and most of the sidewalk, whereas HRDA mistakes one of the traffic signs for a traffic light and most of the sidewalk for road, which would be detrimental for the safety of the pedestrians standing on the sidewalk. On the nighttime image in the second row, CISS accurately segments most of the sidewalk on the right and also successfully segments part of the terrain, even though the latter appears very dark. On the bottom rainy image, HRDA incorrectly segments a green reflection of a traffic light on the road as traffic light, while CISS correctly assigns this reflection to road and also segments the sidewalk on the right much more precisely.

\subsection{Comparison for Zero-Shot Generalization}
\label{sec:experiments:generalization}

\begin{table*}[!tb]
  \caption{\textbf{Comparison of state-of-the-art domain adaptation methods on zero-shot generalization to BDD100K-night.} All methods are trained on Cityscapes$\to$Dark Zurich. Read as Table~\ref{table:cs2dz}.}
  \label{tab:cs2dz:generalization:bdd100kn}
  \centering
  \setlength\tabcolsep{2pt}
  \footnotesize
  \smallskip
  \begin{tabular*}{\linewidth}{l @{\extracolsep{\fill}} cccccccccccccccccccc}
  \toprule
  Method & \ver{road} & \ver{sidew.} & \ver{build.} & \ver{wall} & \ver{fence} & \ver{pole} & \ver{light} & \ver{sign} & \ver{veget.} & \ver{terrain} & \ver{sky} & \ver{person} & \ver{rider} & \ver{car} & \ver{truck} & \ver{bus} & \ver{train} & \ver{motorc.} & \ver{bicycle} & mIoU\\
  \midrule
  GCMA~\cite{GCMA_UIoU:v1} & 85.8 & 48.1 & 64.1 & 1.4 & 16.3 & 30.4 & \textbf{23.7} & 34.9 & 43.1 & 6.8 & 5.9 & 65.4 & \textbf{76.8} & 78.8 & 15.3 & 29.8 & 0.0 & 0.0 & 3.8 & 33.2\\
  MGCDA~\cite{MGCDA_UIoU} & 83.9 & 45.8 & 74.1 & 0.4 & 17.0 & 30.4 & 23.6 & 33.8 & 42.1 & 10.8 & 49.9 & 65.7 & 65.9 & 79.7 & 10.3 & 26.5 & 0.0 & 0.0 & 3.7 & 34.9\\
  DANNet~\cite{DANNet} & 74.1 & 39.9 & 68.3 & 2.6 & 6.1 & 21.3 & 10.6 & 30.6 & 36.3 & 13.4 & 51.8 & 56.0 & 18.7 & 66.6 & 17.6 & 3.0 & 0.0 & 0.0 & 0.8 & 27.2\\
  \midrule
  DAFormer~\cite{daformer:domain:adaptation} & 68.7 & 25.1 & 70.7 & 2.2 & 13.5 & 28.7 & 20.6 & 40.1 & 25.8 & 10.1 & 29.1 & 55.5 & 43.5 & 71.9 & 5.2 & 12.1 & 0.0 & \textbf{0.2} & 3.2 & 27.7 \\
  SePiCo~\cite{sepico:adaptation} & 87.3 & 48.3 & 80.2 & 3.3 & 12.2 & \textbf{37.9} & 20.1 & \textbf{51.4} & \textbf{47.6} & 20.5 & \textbf{65.5} & 67.6 & 67.1 & 83.7 & 29.9 & 46.3 & 0.0 & 0.0 & 1.9 & 40.6\\
  HRDA~\cite{hrda:domain:adaptation} & 84.8 & 49.6 & 77.0 & 4.5 & 26.9 & 35.7 & 21.7 & 47.3 & 35.4 & 12.3 & 60.4 & 66.9 & 27.6 & 81.4 & \textbf{53.1} & 65.2 & 0.0 & 0.0 & 13.9 & 40.2\\
  MIC~\cite{mic:adaptation} & 78.0 & 43.4 & \textbf{80.4} & \textbf{5.6} & 30.5 & 36.6 & 16.6 & 44.6 & 33.0 & 14.5 & 49.8 & 69.1 & 30.1 & 76.5 & 51.3 & \textbf{78.6} & 0.0 & 0.0 & \textbf{14.1} & 39.6 \\
  CISS (ours) & \textbf{88.6} & \textbf{51.3} & 78.4 & \textbf{5.6} & \textbf{34.7} & 37.2 & 19.7 & 44.4 & 32.5 & \textbf{46.5} & 57.9 & \textbf{71.3} & 73.4 & \textbf{84.7} & 39.6 & 18.6 & 0.0 & 0.0 & 10.2 & \textbf{41.8} \\
  \bottomrule
  \end{tabular*}
\end{table*}

\begin{table*}[!tb]
  \caption{\textbf{Comparison of state-of-the-art domain adaptation methods on zero-shot generalization to ACDC-night.} All methods are trained on Cityscapes$\to$Dark Zurich. Read as Table~\ref{table:cs2dz}.}
  \label{table:cs2dz:generalization:acdcn}
  \centering
  \setlength\tabcolsep{2pt}
  \footnotesize
  \smallskip
  \begin{tabular*}{\linewidth}{l @{\extracolsep{\fill}} cccccccccccccccccccc}
  \toprule
  Method & \ver{road} & \ver{sidew.} & \ver{build.} & \ver{wall} & \ver{fence} & \ver{pole} & \ver{light} & \ver{sign} & \ver{veget.} & \ver{terrain} & \ver{sky} & \ver{person} & \ver{rider} & \ver{car} & \ver{truck} & \ver{bus} & \ver{train} & \ver{motorc.} & \ver{bicycle} & mIoU\\
  \midrule
  GCMA~\cite{GCMA_UIoU:v1} & 78.6 & 45.9 & 58.5 & 17.7 & 18.6 & 37.5 & 43.6 & 43.5 & 58.7 & 39.2 & 22.4 & 57.9 & 29.9 & 72.1 & 21.5 & 56.2 & 41.8 & 35.7 & 35.4 & 42.9\\
  MGCDA~\cite{MGCDA_UIoU} & 74.5 & 52.5 & 69.4 & 7.7 & 10.8 & 38.4 & 40.2 & 43.3 & 61.5 & 36.3 & 37.6 & 55.3 & 25.6 & 71.2 & 10.9 & 46.4 & 32.6 & 27.3 & 33.8 & 40.8\\
  DANNet~\cite{DANNet} & 90.7 & 61.1 & 75.5 & 35.9 & 28.8 & 26.6 & 31.4 & 30.6 & \textbf{70.8} & 39.4 & \textbf{78.7} & 49.9 & 28.8 & 65.9 & 24.7 & 44.1 & 61.1 & 25.9 & 34.5 & 47.6\\
  \midrule
  DAFormer~\cite{daformer:domain:adaptation} & 91.5 & 61.9 & 67.7 & 30.9 & 15.0 & 44.6 & 43.3 & 40.0 & 55.2 & 41.4 & 44.6 & 54.1 & 31.9 & 74.7 & 9.1 & 44.8 & 83.3 & 38.1 & 45.0 & 48.3\\
  HRDA~\cite{hrda:domain:adaptation} & 87.5 & 48.1 & 77.6 & 43.2 & 23.2 & 51.1 & \textbf{53.2} & 50.2 & 54.1 & 35.8 & 55.6 & 63.2 & 40.4 & 80.7 & \textbf{63.5} & 81.8 & 80.6 & 46.0 & 49.5 & 57.1\\
  MIC~\cite{mic:adaptation} & \textbf{93.0} & \textbf{68.4} & \textbf{85.1} & \textbf{50.7} & \textbf{32.5} & \textbf{55.2} & 43.2 & \textbf{55.5} & 65.3 & \textbf{50.5} & 66.1 & \textbf{66.9} & \textbf{48.8} & 78.0 & 43.2 & 74.1 & \textbf{89.1} & \textbf{53.4} & \textbf{53.8} & 61.7\\
  CISS (ours) & 92.8 & 67.0 & 83.4 & 49.2 & 21.0 & 51.8 & 42.4 & 55.2 & 69.7 & 46.1 & 76.4 & 66.4 & 42.9 & \textbf{82.3} & 62.9 & \textbf{82.8} & 88.2 & 48.6 & 50.3 & \textbf{62.1}\\
  \bottomrule
  \end{tabular*}%
\end{table*}

To further test the robustness and generality of CISS for condition-level adaptation, we compare it in Tables~\ref{tab:cs2dz:generalization:bdd100kn}
\blue{and \ref{table:cs2dz:generalization:acdcn}}
to state-of-the-art domain adaptation methods on the challenging nighttime sets of BDD100K-night
\blue{and ACDC-night} respectively,
for zero-shot generalization in night time. In particular, all compared models are the same as those which have been evaluated in Table~\ref{table:cs2dz} and they have been trained for adaptation from Cityscapes to Dark Zurich~\cite{MGCDA_UIoU}. These methods are evaluated here on BDD100K-night \blue{(Table~\ref{tab:cs2dz:generalization:bdd100kn}) and ACDC-night (Table~\ref{table:cs2dz:generalization:acdcn})}, which they have not seen at all during training.
CISS outperforms all other methods \blue{both} on BDD100K-night \blue{and ACDC-night}, achieving mean IoU \blue{scores} of 41.8\% \blue{and 62.1\%} and setting the state of the art for UDA methods from day time to night time on \blue{both of these benchmarks at the time of submission.} Note \blue{in particular} that BDD100K-night represents a highly differentiated domain from the original target domain of Dark Zurich in this comparison, as the former set has been recorded in North America while the latter set has been recorded in central Europe. This differentiation makes the examined setting all the more challenging and the top performance of CISS in this setting is all the more significant as evidence for the increased ability of our method to generalize across different sets besides the original target-domain set.

\begin{table*}[!tb]
  \caption{\textbf{Comparison of state-of-the-art unsupervised domain adaptation methods on Cityscapes$\to$ACDC for rain.} The first, second, third and fourth groups of rows present methods trained externally on Cityscapes$\to$Dark Zurich, DeepLabv2-based UDA methods, a DeepLabv3+-based UDA method, and SegFormer-based UDA methods, respectively. Best result per column in bold, second-best underlined.}
  \label{table:cs2acdc:sota:rain}
  \centering
  \setlength\tabcolsep{2pt}
  \footnotesize
  \smallskip
  \begin{tabular*}{\linewidth}{l @{\extracolsep{\fill}} cccccccccccccccccccc}
  \toprule
  Method & \ver{road} & \ver{sidew.} & \ver{build.} & \ver{wall} & \ver{fence} & \ver{pole} & \ver{light} & \ver{sign} & \ver{veget.} & \ver{terrain} & \ver{sky} & \ver{person} & \ver{rider} & \ver{car} & \ver{truck} & \ver{bus} & \ver{train} & \ver{motorc.} & \ver{bicycle} & mIoU\\
  \midrule
  GCMA~\cite{GCMA_UIoU:v1} & 81.1 & 48.0 & 84.8 & 25.0 & 37.3 & 49.8 & 66.5 & 66.2 & 92.1 & 43.5 & 97.6 & 54.5 & 20.4 & 85.5 & 47.3 & 34.6 & 71.3 & 40.3 & 56.7 & 58.0\\
  MGCDA~\cite{MGCDA_UIoU} & 80.5 & 46.5 & 79.9 & 16.0 & 28.8 & 44.9 & 60.0 & 61.5 & 90.3 & 44.8 & 97.1 & 51.1 & 23.1 & 82.3 & 33.4 & 30.2 & 69.1 & 36.5 & 53.8 & 54.2\\
  \midrule
  AdaptSegNet~\cite{adapt:structured:output:cvpr18} & 81.2 & 43.2 & 83.3 & 27.3 & 31.4 & 23.0 & 41.4 & 40.5 & 87.2 & 35.0 & 93.1 & 40.2 & 15.5 & 73.9 & 45.7 & 34.9 & 57.0 & 27.1 & 49.1 & 49.0\\
  BDL~\cite{bidirectional:learning:adaptation} & 79.1 & 39.0 & 82.8 & 30.0 & 34.5 & 28.1 & 40.1 & 47.3 & 87.0 & 28.7 & 91.8 & 40.6 & 17.8 & 74.6 & 46.3 & 36.7 & 60.4 & 33.2 & 46.3 & 49.7\\
  CLAN~\cite{category:adversaries:adaptation} & 77.5 & 40.0 & 46.8 & 24.9 & 30.3 & 28.1 & 37.7 & 48.3 & 83.8 & 37.0 & 6.6 & 45.7 & 17.4 & 79.7 & 43.7 & 42.9 & 63.7 & 35.0 & 46.1 & 44.0\\
  CRST~\cite{crst:adaptation} & 58.8 & 26.4 & 77.1 & 20.0 & 12.1 & 32.8 & 45.3 & 41.7 & 78.6 & 38.4 & 95.7 & 40.5 & 12.8 & 74.7 & 25.6 & 5.5 & 51.8 & 23.7 & 10.9 & 40.6\\
  FDA~\cite{fda:adaptation} & 76.6 & 45.0 & 82.9 & 37.0 & 35.6 & 34.8 & 49.8 & 52.0 & 88.7 & 37.8 & 88.8 & 43.6 & 17.4 & 76.8 & 46.5 & 53.6 & 64.8 & 34.5 & 45.5 & 53.3\\
  SIM~\cite{sim:adaptation} & 76.6 & 29.6 & 85.7 & 20.4 & 28.7 & 21.3 & 37.4 & 34.2 & 87.3 & 34.8 & 94.0 & 29.4 & 16.6 & 73.2 & 46.1 & 22.3 & 46.2 & 21.8 & 39.3 & 44.5\\
  MRNet~\cite{mrnet:rectifying} & 70.5 & 9.9 & 46.5 & 35.6 & 36.1 & 36.5 & 56.4 & 56.2 & 90.2 & 41.3 & 4.3 & 53.0 & 23.5 & 81.6 & 39.3 & 26.7 & 57.8 & 43.6 & 54.5 & 45.4\\
  DACS~\cite{dacs:domain:adaptation} & 69.3 & 41.8 & 84.3 & 30.1 & 20.6 & 38.4 & 38.3 & 54.8 & 83.5 & 38.9 & 82.8 & 41.5 & 14.6 & 76.3 & 47.4 & 30.7 & 53.7 & 30.4 & 49.6 & 48.8\\
  CISS-DeepLabv2 (ours) & 78.6 & 43.4 & 84.9 & 40.2 & 38.7 & 37.4 & 48.9 & 56.9 & 88.3 & 34.2 & 92.5 & 44.9 & 16.9 & 81.0 & 53.0 & 50.7 & 67.2 & 29.9 & 41.8 & 54.2\\
  \midrule
  MALL~\cite{mall:domain:generalization} & 75.9 & 38.1 & 87.6 & 35.9 & 38.6 & 45.9 & 60.7 & 60.1 & 88.8 & 38.7 & 96.6 & 48.9 & 14.2 & 84.8 & 56.4 & 63.8 & 71.7 & 27.7 & 47.8 & 57.0\\
  \midrule
  DAFormer~\cite{daformer:domain:adaptation} & 73.1 & 46.7 & 92.2 & 55.9 & 40.5 & 54.9 & 65.6 & 64.9 & 93.1 & 40.8 & 89.8 & 58.5 & 20.6 & 86.1 & 63.5 & 66.4 & 83.0 & 46.6 & 53.4 & 62.9\\
  SePiCo~\cite{sepico:adaptation} & 80.1 & 47.3 & 90.1 & 48.9 & \underline{48.2} & 57.0 & 70.4 & 66.5 & 93.2 & 43.2 & 93.8 & 67.3 & 26.4 & 89.0 & 68.1 & 71.5 & 88.8 & 49.6 & 57.0 & 66.1\\
  HRDA~\cite{hrda:domain:adaptation} & \underline{92.4} & \underline{73.6} & 93.8 & \underline{67.0} & 46.3 & \textbf{63.0} & 74.5 & 74.2 & \underline{93.7} & 46.1 & 97.6 & 69.4 & \underline{32.5} & 91.7 & \textbf{79.9} & \underline{90.5} & \underline{89.0} & 57.8 & \underline{66.0} & 73.6\\
  MIC~\cite{mic:adaptation} & 90.6 & 69.7 & \underline{93.9} & 61.0 & 47.5 & \underline{62.9} & \textbf{75.3} & \textbf{75.1} & \underline{93.7} & \textbf{48.3} & \textbf{98.2} & \textbf{72.0} & 31.4 & \textbf{92.7} & \underline{78.4} & \textbf{93.1} & \textbf{89.9} & \textbf{61.7} & 65.4 & \underline{73.7} \\
  CISS (ours) & \textbf{92.5} & \textbf{74.7} & \textbf{94.8} & \textbf{70.3} & \textbf{49.5} & 61.2 & \underline{74.8} & \underline{74.5} & \textbf{94.0} & \underline{47.9} & \textbf{98.2} & \underline{70.5} & \textbf{37.8} & \underline{92.1} & 75.0 & 89.8 & 88.0 & \underline{59.2} & \textbf{66.1} & \textbf{74.3} \\
  \bottomrule
  \end{tabular*}%
\end{table*}

\begin{table*}[!tb]
  \caption{\textbf{Comparison of state-of-the-art unsupervised domain adaptation methods on Cityscapes$\to$ACDC for night time.} The first, second, third and fourth groups of rows present methods trained externally on Cityscapes$\to$Dark Zurich, DeepLabv2-based UDA methods, a DeepLabv3+-based UDA method, and SegFormer-based UDA methods, respectively. Best result per column in bold, second-best underlined.}
  \label{table:cs2acdc:sota:night}
  \centering
  \setlength\tabcolsep{2pt}
  \footnotesize
  \smallskip
  \begin{tabular*}{\linewidth}{l @{\extracolsep{\fill}} cccccccccccccccccccc}
  \toprule
  Method & \ver{road} & \ver{sidew.} & \ver{build.} & \ver{wall} & \ver{fence} & \ver{pole} & \ver{light} & \ver{sign} & \ver{veget.} & \ver{terrain} & \ver{sky} & \ver{person} & \ver{rider} & \ver{car} & \ver{truck} & \ver{bus} & \ver{train} & \ver{motorc.} & \ver{bicycle} & mIoU\\
  \midrule
  GCMA~\cite{GCMA_UIoU:v1} & 78.6 & 45.9 & 58.5 & 17.7 & 18.6 & 37.5 & \underline{43.6} & 43.5 & 58.7 & 39.2 & 22.4 & 57.9 & 29.9 & 72.1 & 21.5 & 56.2 & 41.8 & 35.7 & 35.4 & 42.9\\
  MGCDA~\cite{MGCDA_UIoU} & 74.5 & 52.5 & 69.4 & 7.7 & 10.8 & 38.4 & 40.2 & 43.3 & \underline{61.5} & 36.3 & \underline{37.6} & 55.3 & 25.6 & 71.2 & 10.9 & 46.4 & 32.6 & 27.3 & 33.8 & 40.8\\
  DANNet~\cite{DANNet} & 90.7 & 61.1 & 75.5 & 35.9 & \underline{28.8} & 26.6 & 31.4 & 30.6 & \textbf{70.8} & 39.4 & \textbf{78.7} & 49.9 & 28.8 & 65.9 & 24.7 & 44.1 & 61.1 & 25.9 & 34.5 & 47.6\\
  \midrule
  AdaptSegNet~\cite{adapt:structured:output:cvpr18} & 84.9 & 39.9 & 66.8 & 17.2 & 17.7 & 13.4 & 17.6 & 16.4 & 39.6 & 16.1 & 5.7 & 42.8 & 21.4 & 44.8 & 11.9 & 13.0 & 39.1 & 27.5 & 28.4 & 29.7\\
  BDL~\cite{bidirectional:learning:adaptation} & 87.1 & 49.6 & 68.8 & 20.2 & 17.5 & 16.7 & 19.9 & 24.1 & 39.1 & 23.7 & 0.2 & 42.0 & 20.4 & 63.7 & 18.0 & 27.0 & 45.6 & 27.8 & 31.3 & 33.8\\
  CLAN~\cite{category:adversaries:adaptation} & 82.3 & 28.8 & 65.9 & 15.1 & 9.3 & 22.1 & 16.1 & 26.5 & 39.2 & 23.4 & 0.4 & 45.9 & 25.4 & 63.6 & 9.5 & 24.2 & 39.8 & 31.5 & 31.1 & 31.6\\
  CRST~\cite{crst:adaptation} & 43.9 & 10.0 & 57.3 & 10.0 & 5.1 & 29.3 & 27.0 & 18.6 & 6.9 & 8.2 & 0.3 & 36.9 & 17.9 & 48.5 & 4.9 & 1.8 & 29.4 & 7.3 & 8.8 & 19.6\\
  FDA~\cite{fda:adaptation} & 82.7 & 39.4 & 57.0 & 14.7 & 7.6 & 26.1 & 37.8 & 30.5 & 53.2 & 14.0 & 15.3 & 48.0 & 28.8 & 62.6 & 26.6 & 47.5 & 51.5 & 27.0 & 35.0 & 37.1\\
  SIM~\cite{sim:adaptation} & 87.0 & 48.4 & 42.1 & 6.3 & 8.3 & 15.8 & 8.4 & 17.6 & 21.7 & 22.8 & 0.1 & 39.3 & 22.1 & 60.3 & 8.7 & 18.2 & 42.3 & 30.1 & 32.9 & 28.0\\
  MRNet~\cite{mrnet:rectifying} & 83.6 & 36.3 & 65.6 & 8.1 & 8.2 & 21.5 & 30.0 & 23.7 & 39.4 & 24.2 & 0.0 & 44.1 & 26.0 & 64.9 & 0.8 & 3.6 & 7.6 & 10.3 & 31.8 & 27.9\\
  DACS~\cite{dacs:domain:adaptation} & 84.8 & 52.5 & 64.8 & 17.5 & 16.0 & 30.5 & 25.1 & 33.9 & 38.4 & 10.7 & 2.7 & 40.7 & 21.2 & 63.9 & 16.4 & 36.6 & 45.4 & 19.5 & 23.4 & 33.9\\
  CISS-DeepLabv2 (ours) & 77.5 & 29.6 & 59.3 & 18.0 & 14.0 & 31.0 & 39.3 & 35.6 & 41.5 & 12.8 & 2.1 & 48.6 & 31.7 & 69.1 & 26.8 & 60.9 & 53.0 & 23.6 & 34.7 & 37.3\\
  \midrule
  MALL~\cite{mall:domain:generalization} & 78.9 & 26.8 & 62.2 & 25.3 & 19.9 & 32.3 & 32.6 & 31.4 & 49.9 & 27.9 & 13.5 & 47.3 & 19.6 & 61.0 & 19.2 & 35.4 & 56.0 & 29.7 & 31.4 & 36.9\\
  \midrule
  DAFormer~\cite{daformer:domain:adaptation} & 92.3 & 64.6 & 70.1 & 28.7 & 18.5 & 45.8 & 11.3 & 41.5 & 42.7 & 41.9 & 0.0 & 55.4 & 29.8 & 74.3 & 40.3 & 45.8 & 81.3 & 39.4 & 47.0 & 45.8\\
  SePiCo~\cite{sepico:adaptation} & 89.9 & 56.8 & 75.6 & 35.3 & 28.4 & 49.5 & 24.7 & 50.1 & 43.4 & 44.5 & 4.8 & 61.1 & 34.1 & 77.3 & 62.0 & 52.9 & 79.5 & 41.2 & 48.3 & 50.5\\
  HRDA~\cite{hrda:domain:adaptation} & 87.2 & 46.9 & 79.1 & 46.2 & 18.0 & 51.4 & 41.0 & 48.5 & 41.8 & 46.7 & 0.0 & 63.2 & 36.9 & 81.0 & \underline{65.2} & \underline{77.7} & 83.6 & 46.0 & 49.0 & 53.1\\
  MIC~\cite{mic:adaptation} & \textbf{95.5} & \textbf{78.0} & \textbf{82.1} & \textbf{49.1} & \textbf{36.4} & \textbf{53.1} & 40.6 & \textbf{61.7} & 44.2 & \underline{51.4} & 8.3 & \textbf{66.4} & \textbf{45.1} & \underline{83.6} & \textbf{68.5} & \textbf{82.5} & \textbf{89.0} & \underline{52.3} & \textbf{54.5} & \textbf{60.1} \\
  CISS (ours) & \underline{94.7} & \underline{74.5} & \underline{81.2} & \underline{48.2} & 28.4 & \underline{52.2} & \textbf{50.1} & \underline{58.6} & 43.2 & \textbf{53.4} & 2.6 & \underline{65.7} & \underline{39.0} & \textbf{83.8} & 63.2 & 74.7 & \underline{86.6} & \textbf{52.9} & \underline{53.5} & \underline{58.2}\\
  \bottomrule
  \end{tabular*}%
\end{table*}

\begin{table*}[!tb]
  \caption{\textbf{Comparison of state-of-the-art unsupervised domain adaptation methods on Cityscapes$\to$ACDC for snow.} The first, second, third and fourth groups of rows present methods trained externally on Cityscapes$\to$Dark Zurich, DeepLabv2-based UDA methods, a DeepLabv3+-based UDA method, and SegFormer-based UDA methods, respectively. Best result per column in bold.}
  \label{table:cs2acdc:sota:snow}
  \centering
  \setlength\tabcolsep{2pt}
  \footnotesize
  \smallskip
  \begin{tabular*}{\linewidth}{l @{\extracolsep{\fill}} cccccccccccccccccccc}
  \toprule
  Method & \ver{road} & \ver{sidew.} & \ver{build.} & \ver{wall} & \ver{fence} & \ver{pole} & \ver{light} & \ver{sign} & \ver{veget.} & \ver{terrain} & \ver{sky} & \ver{person} & \ver{rider} & \ver{car} & \ver{truck} & \ver{bus} & \ver{train} & \ver{motorc.} & \ver{bicycle} & mIoU\\
  \midrule
  GCMA~\cite{GCMA_UIoU:v1} & 79.7 & 49.5 & 75.3 & 17.5 & 37.9 & 43.2 & 59.0 & 61.9 & 78.8 & 2.2 & 95.5 & 62.5 & 33.6 & 83.2 & 42.5 & 43.4 & 72.1 & 32.2 & 51.1 & 53.7\\
  MGCDA~\cite{MGCDA_UIoU} & 80.1 & 49.5 & 70.2 & 6.1 & 27.8 & 39.6 & 55.4 & 58.0 & 76.0 & 0.3 & 95.5 & 57.5 & 35.7 & 81.0 & 28.6 & 48.9 & 70.3 & 27.8 & 50.5 & 50.5\\
  \midrule
  AdaptSegNet~\cite{adapt:structured:output:cvpr18} & 51.3 & 32.5 & 47.3 & 21.5 & 31.5 & 13.2 & 37.8 & 23.2 & 76.0 & 2.6 & 4.5 & 49.9 & 23.1 & 68.7 & 38.3 & 31.8 & 51.5 & 21.7 & 45.0 & 35.3\\
  BDL~\cite{bidirectional:learning:adaptation} & 42.3 & 36.4 & 60.2 & 15.7 & 30.4 & 15.1 & 41.4 & 30.4 & 71.3 & 1.7 & 11.2 & 46.8 & 27.8 & 57.7 & 38.6 & 34.1 & 59.2 & 28.1 & 43.7 & 36.4\\
  CLAN~\cite{category:adversaries:adaptation} & 71.8 & 26.0 & 37.3 & 12.5 & 27.0 & 21.1 & 32.0 & 41.1 & 78.5 & 1.9 & 0.9 & 50.9 & 23.9 & 82.4 & 43.2 & 39.5 & 61.6 & 25.2 & 39.4 & 37.7\\
  CRST~\cite{crst:adaptation} & 63.5 & 38.2 & 66.8 & 12.8 & 9.2 & 29.0 & 44.8 & 40.3 & 68.5 & 0.8 & 65.1 & 44.6 & 23.8 & 70.0 & 1.2 & 19.0 & 39.1 & 11.4 & 6.0 & 34.4\\
  FDA~\cite{fda:adaptation} & 74.6 & 30.9 & 56.1 & 20.5 & 34.8 & 28.7 & 53.9 & 47.8 & 80.5 & 1.1 & 55.9 & 53.1 & 37.9 & 79.7 & 40.5 & 51.9 & 67.4 & 34.3 & 41.8 & 46.9\\
  SIM~\cite{sim:adaptation} & 72.1 & 26.7 & 39.4 & 13.3 & 29.5 & 15.3 & 26.4 & 17.9 & 76.4 & 4.8 & 5.1 & 45.9 & 32.0 & 76.2 & 29.8 & 26.6 & 48.3 & 23.2 & 24.2 & 33.3\\
  MRNet~\cite{mrnet:rectifying} & 67.7 & 3.5 & 36.8 & 8.3 & 24.8 & 18.0 & 52.6 & 55.4 & 82.4 & 0.5 & 0.1 & 62.2 & 30.2 & 79.2 & 32.1 & 59.3 & 58.4 & 29.1 & 35.8 & 38.7\\
  DACS~\cite{dacs:domain:adaptation} & 52.4 & 13.7 & 77.7 & 14.2 & 24.7 & 33.2 & 40.3 & 50.6 & 78.8 & 0.8 & 34.2 & 51.7 & 22.2 & 75.0 & 30.8 & 30.6 & 58.4 & 19.8 & 43.9 & 39.6\\
  CISS-DeepLabv2 (ours) & 75.5 & 39.3 & 67.9 & 29.8 & 37.9 & 31.1 & 49.6 & 54.0 & 79.5 & 1.6 & 77.2 & 53.7 & 43.5 & 81.5 & 41.5 & 37.2 & 69.1 & 22.7 & 41.2 & 49.1\\
  \midrule
  MALL~\cite{mall:domain:generalization} & 78.2 & 40.9 & 78.8 & 19.1 & 36.6 & 39.7 & 60.9 & 51.6 & 80.9 & 6.8 & 90.5 & 54.8 & 28.1 & 82.9 & 40.3 & 58.6 & 68.4 & 13.4 & 46.6 & 51.4\\
  \midrule
  DAFormer~\cite{daformer:domain:adaptation} & 38.1 & 41.3 & 88.3 & 42.1 & 47.2 & 54.2 & 71.1 & 64.2 & 91.2 & 4.5 & 32.8 & 66.0 & 36.4 & 88.0 & 54.4 & 71.3 & 84.5 & 46.0 & 54.8 & 56.7\\
  SePiCo~\cite{sepico:adaptation} & 40.5 & 33.7 & 87.1 & 29.2 & 50.0 & 57.6 & 76.1 & 66.1 & 90.4 & 4.2 & 42.8 & 71.9 & 41.5 & 89.3 & 66.4 & 69.7 & 88.6 & 37.2 & 57.8 & 57.9\\
  HRDA~\cite{hrda:domain:adaptation} & 82.5 & 45.5 & 90.4 & 55.3 & 49.9 & 58.9 & 77.7 & \textbf{71.9} & 91.3 & 6.0 & 96.2 & 79.6 & 62.8 & 92.0 & \textbf{73.8} & 73.1 & \textbf{90.4} & 52.0 & 70.7 & 69.5\\
  MIC~\cite{mic:adaptation} & 79.3 & 36.0 & 90.9 & 55.0 & 48.6 & \textbf{59.6} & \textbf{79.4} & 70.6 & \textbf{91.8} & \textbf{8.8} & \textbf{96.8} & \textbf{80.8} & \textbf{63.5} & \textbf{92.5} & \textbf{73.8} & \textbf{80.4} & 88.8 & \textbf{54.0} & \textbf{75.0} & \textbf{69.8} \\
  CISS (ours) & \textbf{84.1} & \textbf{51.1} & \textbf{91.0} & \textbf{58.6} & \textbf{50.5} & 58.0 & 77.5 & 70.4 & 91.3 & 4.7 & 96.7 & 78.8 & 60.3 & 91.6 & 71.0 & 79.1 & 87.0 & 51.0 & 70.5 & 69.6 \\
  \bottomrule
  \end{tabular*}%
\end{table*}

\begin{table*}[!tb]
  \caption{\textbf{Comparison of state-of-the-art unsupervised domain adaptation methods on Cityscapes$\to$ACDC for fog.} The first, second, third, fourth and fifth groups of rows present methods trained externally on Cityscapes$\to$Dark Zurich, a method trained externally on Cityscapes$\to$Foggy Zurich~\cite{SynRealDataFogECCV18}, DeepLabv2-based UDA methods, a DeepLabv3+-based UDA method, and SegFormer-based UDA methods, respectively. Best result per column in bold, second-best underlined.}
  \label{table:cs2acdc:sota:fog}
  \centering
  \setlength\tabcolsep{2pt}
  \footnotesize
  \smallskip
  \begin{tabular*}{\linewidth}{l @{\extracolsep{\fill}} cccccccccccccccccccc}
  \toprule
  Method & \ver{road} & \ver{sidew.} & \ver{build.} & \ver{wall} & \ver{fence} & \ver{pole} & \ver{light} & \ver{sign} & \ver{veget.} & \ver{terrain} & \ver{sky} & \ver{person} & \ver{rider} & \ver{car} & \ver{truck} & \ver{bus} & \ver{train} & \ver{motorc.} & \ver{bicycle} & mIoU\\
  \midrule
  GCMA~\cite{GCMA_UIoU:v1} & 80.8 & 53.5 & 70.1 & 29.2 & 20.7 & 38.4 & 53.0 & 60.9 & 70.2 & 46.5 & 95.4 & 44.2 & 38.0 & 76.6 & 52.4 & 49.7 & 56.8 & 41.0 & 17.6 & 52.4\\
  MGCDA~\cite{MGCDA_UIoU} & 71.7 & 47.3 & 65.7 & 18.2 & 15.3 & 34.4 & 48.6 & 59.9 & 64.9 & 24.7 & 95.4 & 44.8 & 23.8 & 73.3 & 36.1 & 45.4 & 63.9 & 23.9 & 15.4 & 45.9\\
  \midrule
  CuDA-Net~\cite{cudanet:fog:domain:adaptation} & 83.2 & 45.9 & 81.7 & 35.5 & 22.7 & 40.7 & 55.5 & 55.6 & 81.1 & 63.8 & 95.6 & 45.2 & 24.9 & 78.7 & 41.1 & 48.3 & 77.8 & 52.0 & 27.1 & 55.6\\
  \midrule
  AdaptSegNet~\cite{adapt:structured:output:cvpr18} & 35.4 & 45.9 & 35.4 & 25.6 & 17.5 & 9.0 & 32.5 & 23.1 & 70.5 & 47.4 & 11.6 & 22.3 & 28.2 & 44.4 & 43.9 & 35.0 & 46.0 & 15.6 & 15.0 & 31.8\\
  BDL~\cite{bidirectional:learning:adaptation} & 36.9 & 37.8 & 47.0 & 28.2 & 21.6 & 13.7 & 37.2 & 34.5 & 67.2 & 49.4 & 27.6 & 29.1 & 51.3 & 58.5 & 49.4 & 51.8 & 30.3 & 21.4 & 22.5 & 37.7\\
  CLAN~\cite{category:adversaries:adaptation} & 48.8 & 41.3 & 29.6 & 27.2 & 21.0 & 16.1 & 41.1 & 39.6 & 67.7 & 50.2 & 15.4 & 36.2 & 30.8 & 72.2 & 52.2 & 54.4 & 47.2 & 27.1 & 22.6 & 39.0\\
  CRST~\cite{crst:adaptation} & 59.7 & 29.6 & 70.9 & 11.3 & 11.4 & 29.9 & 41.4 & 38.6 & 61.7 & 31.6 & 96.6 & 36.0 & 7.9 & 62.4 & 19.7 & 4.6 & 49.4 & 9.0 & 7.6 & 35.8\\
  FDA~\cite{fda:adaptation} & 68.8 & 37.3 & 27.1 & 27.6 & 19.8 & 21.6 & 37.5 & 43.3 & 74.9 & 43.7 & 33.1 & 35.0 & 21.5 & 65.7 & 44.6 & 45.3 & 47.1 & 41.5 & 15.8 & 39.5\\
  SIM~\cite{sim:adaptation} & 76.7 & 43.1 & 23.5 & 23.6 & 17.9 & 10.9 & 32.1 & 15.3 & 70.4 & 50.5 & 21.4 & 34.8 & 44.3 & 58.4 & 50.5 & 55.2 & 34.7 & 23.0 & 8.8 & 36.6\\
  MRNet~\cite{mrnet:rectifying} & 78.6 & 26.1 & 19.6 & 29.0 & 13.5 & 12.0 & 41.9 & 49.0 & 78.2 & 59.0 & 6.6 & 39.8 & 26.1 & 72.5 & 44.8 & 37.9 & 59.6 & 19.1 & 24.1 & 38.8\\
  DACS~\cite{dacs:domain:adaptation} & 34.9 & 51.8 & 79.0 & 22.8 & 24.8 & 22.9 & 20.0 & 46.6 & 50.5 & 50.8 & 19.7 & 38.2 & 25.9 & 69.5 & 44.1 & 48.5 & 29.9 & 28.8 & 16.0 & 38.1\\
  CISS-DeepLabv2 (ours) & 51.7 & 36.9 & 53.4 & 29.6 & 22.1 & 25.3 & 41.3 & 49.2 & 75.8 & 30.8 & 61.6 & 36.2 & 34.6 & 67.3 & 44.5 & 29.7 & 52.6 & 38.7 & 19.0 & 42.1\\
  \midrule
  MALL~\cite{mall:domain:generalization} & 63.7 & 54.3 & 79.8 & 34.8 & 27.4 & 37.9 & 49.1 & 52.6 & 74.9 & 59.6 & 92.9 & 40.2 & 39.0 & 75.4 & 53.0 & 36.4 & 76.4 & 26.8 & 21.5 & 52.4\\
  \midrule
  DAFormer~\cite{daformer:domain:adaptation} & 38.9 & 42.4 & 86.8 & 52.5 & 26.8 & 46.7 & 45.6 & 57.3 & 86.4 & 64.7 & 56.5 & 37.6 & 53.3 & 76.2 & 60.8 & 32.4 & 64.0 & 52.1 & 29.6 & 53.2\\
  SePiCo~\cite{sepico:adaptation} & 42.6 & 51.5 & 87.6 & 51.2 & \textbf{31.2} & \textbf{52.4} & 51.0 & 59.0 & 85.3 & 65.9 & 61.3 & 51.4 & \underline{62.2} & 78.0 & 64.5 & 42.3 & 83.5 & 58.0 & 32.6 & 58.5\\
  HRDA~\cite{hrda:domain:adaptation} & 93.0 & 73.5 & 89.1 & \textbf{56.4} & 27.3 & 51.2 & \textbf{62.2} & \textbf{69.5} & 86.5 & 70.3 & \underline{98.0} & 53.4 & 61.9 & \underline{85.6} & \textbf{77.1} & \underline{88.3} & \underline{84.9} & \textbf{64.1} & 36.6 & \underline{69.9} \\
  MIC~\cite{mic:adaptation} & \textbf{94.5} & \textbf{78.6} & \underline{89.5} & \underline{55.4} & 27.8 & \underline{51.7} & \underline{60.9} & \underline{65.7} & \textbf{87.8} & \textbf{75.3} & \textbf{98.1} & \textbf{55.4} & 62.0 & \textbf{86.6} & 75.6 & \textbf{92.1} & \textbf{89.2} & \underline{62.8} & \textbf{42.6} & \textbf{71.1} \\
  CISS (ours) & \underline{94.1} & \underline{76.2} & \textbf{89.8} & 55.1 & \underline{29.6} & 50.3 & \underline{61.4} & 65.4 & \underline{87.0} & \underline{72.0} & 97.9 & \underline{55.2} & \textbf{62.9} & 83.7 & \underline{75.9} & 60.6 & 83.0 & 62.1 & \textbf{42.6} & 68.7 \\
  \bottomrule
  \end{tabular*}%
\end{table*}

Let us focus on the comparison between CISS and its most direct competitors, HRDA and MIC, on \blue{these two} generalization \blue{experiments on BDD100K-night and ACDC-night (cf.\ the last three rows of Tables~\ref{tab:cs2dz:generalization:bdd100kn} and \ref{table:cs2dz:generalization:acdcn}), respectively.} Recall that the fully-fledged models of both MIC and CISS are implemented on top of HRDA, so the latter is effectively the common baseline of the two former. \blue{On BDD100K-night}, we observe that MIC (39.6\% mean IoU) performs worse than the common baseline, HRDA (40.2\% mean IoU), in this generalization setting, even though the examined MIC model outperforms the examined HRDA model substantially on the original test set of Dark Zurich (cf.\ Table~\ref{table:cs2dz}). This evidences that MIC has fitted more tightly to the particular target-domain set on which it has been trained, i.e.\ Dark Zurich, than the HRDA baseline, which impairs the generalization of MIC on BDD100K-night. By contrast, CISS (41.8\% mean IoU) outperforms significantly the common baseline, HRDA, in this generalization experiment. That is, compared to their common HRDA baseline, MIC performs worse while CISS performs significantly better. \blue{At the same time, on ACDC-night, which represents a domain that is closer to the training-time target domain of Dark Zurich than BDD100K-night---as ACDC and Dark Zurich have been captured in the same geographic region, CISS (62.1\% mean IoU) still substantially outperforms HRDA (57.1\% mean IoU). CISS also outperforms MIC (61.7\% mean IoU) on ACDC-night and it achieves a more stable performance across all individual classes than MIC, similarly to the respective finding we had in Sec.~\ref{sec:experiments:sota:cs2dz}.} We thus draw the conclusion that our feature invariance loss enables CISS to learn more general features than both HRDA and MIC under the large domain shift between day time and night time, granting our model a better ability to generalize to target nighttime sets that are unseen during training.

\subsection{Comparison on Individual Conditions of ACDC}
\label{sec:experiments:acdc:individual}

In this section, we provide a condition-specific comparison of CISS to competing domain adaptation methods on the four adverse conditions of ACDC. More specifically, in Tables~\ref{table:cs2acdc:sota:rain}, \ref{table:cs2acdc:sota:night}, \ref{table:cs2acdc:sota:snow}, and \ref{table:cs2acdc:sota:fog}, we provide detailed class-level IoU results as well as mean IoU results on the four condition-specific splits of the test set of ACDC, i.e.\ the rain, night time, snow, and fog split respectively, for all methods which have been presented in the comparison of Table~\ref{table:cs2acdc:sota:all} for the entire test set of ACDC. For each of these methods, a single model is trained using the entire training set of ACDC as the target set---the same model which has been evaluated on the entire test set in Table~\ref{table:cs2acdc:sota:all}---and is then evaluated separately on each condition. \blue{Note that the models that are evaluated in this experiment are generally different from those evaluated in Tables~\ref{table:cs2dz}, \ref{tab:cs2dz:generalization:bdd100kn}, and \ref{table:cs2dz:generalization:acdcn}, as the latter set of models is rather trained with Dark Zurich as the target set. Thus, the comparative performance for a pair of methods may differ between the two settings.} In addition, we evaluate DANNet~\cite{DANNet} and CuDA-Net~\cite{cudanet:fog:domain:adaptation}, which are specifically designed for night and fog respectively, so we only report the results which have been originally presented by these works on their respective condition of focus. Finally, we also compare with MALL~\cite{mall:domain:generalization} on all four condition-specific splits; the reason we have not included this method in the comparison on the entire test set in Table~\ref{table:cs2acdc:sota:all} is that the authors do not present the result on the entire test set in their paper and the respective model is not publicly available, which would allow us to reproduce and evaluate the predictions of MALL on the entire test set.

CISS performs favorably compared to other methods on all four conditions of ACDC. In particular, among all methods our method is ranked \emph{first} on rain, second on night time and snow, and third on fog.

CISS achieves the top rank on the rain benchmark of ACDC among all published UDA methods\footnote{\url{https://acdc.vision.ee.ethz.ch/benchmarks\#semanticSegmentation}} (cf.\ Table~\ref{table:cs2acdc:sota:rain}). The performance of CISS on the rain test set of ACDC across the 19 individual classes is consistently excellent, as it is ranked first in 9/19 classes and second in 6/19 classes. Three out of the four remaining classes, i.e.\ truck, bus, and train, are classes with large instances which only appear rarely in the scenes of ACDC~\cite{ACDC} and may thus have large variance in their respective IoU scores.

CISS and MIC dominate the challenging nighttime benchmark of ACDC (cf.\ Table~\ref{table:cs2acdc:sota:night}), scoring 5.1\% and 7.0\% higher on mean IoU respectively than their common baseline, HRDA, which is the third-best method. CISS is ranked first or second among all methods in 14/19 classes and is slightly outperformed by MIC in mean IoU by 1.9\%. Note that CISS is nonetheless better than MIC on multiple classes, notably on hard ones at night time such as traffic light (by 9.5\%) and car. Both of these classes are central for autonomous car perception and undergo a large and thus challenging shift in appearance from the source daytime domain to the target nighttime domain, which involves the activation (car) or relative intensification (traffic light) of light sources and makes these classes harder to distinguish from each other and from other objects with lights at night time, such as buildings and street lights. Another interesting finding of this nighttime evaluation is that although CISS and MIC are overall the top-performing methods, they are both outperformed substantially on vegetation and sky by methods trained specifically on nighttime sets and using weak supervision in the form of cross-time-of-day correspondences~\cite{GCMA_UIoU:v1,MGCDA_UIoU,DANNet}. Vegetation and sky are usually adjacent in ACDC and they both appear very dark in nighttime images, which makes them hard to distinguish from one another at night time and apparently still presents a challenge for completely unsupervised domain adaptation methods trained on Cityscapes$\to$ACDC which needs to be addressed in future work.

CISS is ranked second among all methods on the snowy test set of ACDC (cf.\ Table~\ref{table:cs2acdc:sota:snow}), being marginally outperformed by MIC in mean IoU (by 0.2\%). However, CISS is the top-performing method on the highly challenging classes of both road and sidewalk, which are hardest to segment in snow compared to other adverse conditions~\cite{ACDC} due to snow cover on the ground. In particular, CISS outperforms MIC by 4.8\% on road and by a substantial 15.1\% on sidewalk, with MIC even falling behind HRDA in these classes.

Finally, although CISS is ranked third in mean IoU on the fog benchmark of ACDC (cf.\ Table~\ref{table:cs2acdc:sota:fog}) behind MIC and HRDA, the three methods are comparable with regard to class-level IoU scores. In particular, CISS outperforms HRDA on 9/19 classes and MIC on 6/19 classes, and it is ranked first or second among all methods in 11/19 classes. The higher mean IoU scores of HRDA and MIC compared to CISS are primarily due to the large difference between the IoUs of the two former methods and that of CISS on bus, which is a very rare class in ACDC~\cite{ACDC}.

\begin{table}[!tb]
    \caption{\textbf{Ablation study of CISS on Cityscapes$\to$ACDC.} Evaluation is performed on the validation set of ACDC. ``CE'': cross-entropy loss, ``Inv'': feature invariance loss, ``orig'': original images from respective domain, ``stylized'': images from respective domain stylized with FDA. Mean and standard deviation across three runs are reported.}
    \label{tab:ablation:cs2acdc}
    \resizebox{\columnwidth}{!}{%
    \centering%
    \renewcommand{\arraystretch}{1.1}%
    \setlength\tabcolsep{4pt}%
    \smallskip%
    \small%
    \begin{tabular}{ccccccccc}
        \toprule
        & \multicolumn{3}{c}{\textbf{Source}} && \multicolumn{3}{c}{\textbf{Target}} & mIoU\\
        \cmidrule{2-4} \cmidrule{6-8}
        & CE orig
        & CE stylized
        & Inv
        && CE orig
        & CE stylized
        & Inv
        & \\
        \midrule
        1 & \checkmark & & && \checkmark & & & 64.1$\pm$2.0 \\
        2 & & \checkmark & && \checkmark & & & 65.7$\pm$1.1 \\
        3 & \checkmark & \checkmark & && \checkmark & & & 65.1$\pm$0.7 \\
        4 & \checkmark & & \checkmark && \checkmark & & & 66.6$\pm$0.8 \\
        \midrule
        5 & \checkmark & & && \checkmark & & \checkmark & 66.9$\pm$0.5 \\
        6 & \checkmark & & \checkmark && \checkmark & \checkmark & & 65.7$\pm$1.2 \\
        7 & \checkmark & \checkmark & \checkmark && \checkmark & \checkmark & \checkmark & 68.0$\pm$0.8 \\
        8 (CISS) & \checkmark & & \checkmark && \checkmark & & \checkmark & \textbf{68.2}$\pm$0.4 \\
        \bottomrule
    \end{tabular}}
\end{table}

\subsection{Analysis and Ablation Studies}
\label{sec:experiments:ablations}

\subsubsection{Ablation of Feature Invariance Losses}
\label{sec:experiments:ablations:invariance:loss}

In Table~\ref{tab:ablation:cs2acdc}, we conduct an ablation study of our method w.r.t.\ the various loss terms that are included in our overall loss $\mathcal{L}_{\text{CISS}}$ in \eqref{eq:loss:ciss} and the alternative loss terms that are included in the baseline formulations of \eqref{eq:loss:basic}, \eqref{eq:loss:fda}, and \eqref{eq:loss:ce:full}. Our goal is to demonstrate the benefit of applying our feature invariance loss compared to merely using cross entropy losses on original images as well as to additionally applying cross entropy losses on stylized images, and this both for source-domain and target-domain images. The basic UDA formulation of \eqref{eq:loss:basic}, i.e., plain HRDA, corresponds to row~1. Switching to the FDA loss of \eqref{eq:loss:fda} in row~2, i.e., pixel-level adaptation, improves upon the basic formulation. However, applying cross-entropy loss both for the original source images and their stylized versions (row~3), in the direction of \eqref{eq:loss:ce:full}, does not provide any gain over the FDA loss, evidencing that simultaneous output supervision on different views of images alone is not sufficient for aligning their features. On the contrary, applying the feature invariance loss on the source domain alone (row~4) improves upon the FDA setting of row~2, showing the utility of feature-level adaptation achieved with CISS on top of the pixel-level adaptation with FDA. In addition, the feature invariance loss applied solely on the target domain (row~5) also improves significantly upon the basic UDA setup of row~1. While using stylized target images for applying an additional cross-entropy loss on the target domain hurts performance (cf.\ rows~4 and 6), combining the two feature invariance losses from the source and the target domain in the complete formulation of CISS \eqref{eq:loss:ciss} \blue{(row~8)} improves further compared to applying each of the two losses alone (rows~4 and 5), showing that the two losses synergize and achieve the best result when applied jointly. \blue{In order to further evidence the sufficiency of our feature invariance loss for feature alignment, we additionally evaluate in row~7 a model including all three examined losses, i.e.\ (i) cross-entropy loss on the original images and (ii) on the stylized images as well as (iii) our feature invariance loss, both for the source domain and for the target domain. Compared to our proposed CISS formulation, this model additionally includes cross-entropy losses on the outputs corresponding to the stylized images, however, this inclusion does not provide extra benefit in terms of performance, as our feature invariance loss represents a stronger constraint, applied already at the internal features of the network and explicitly aligning such features across domains.}

\begin{figure}
    \centering
    \includegraphics[width=\linewidth,trim={3cm 3.4cm 2.3cm 3.2cm},clip]{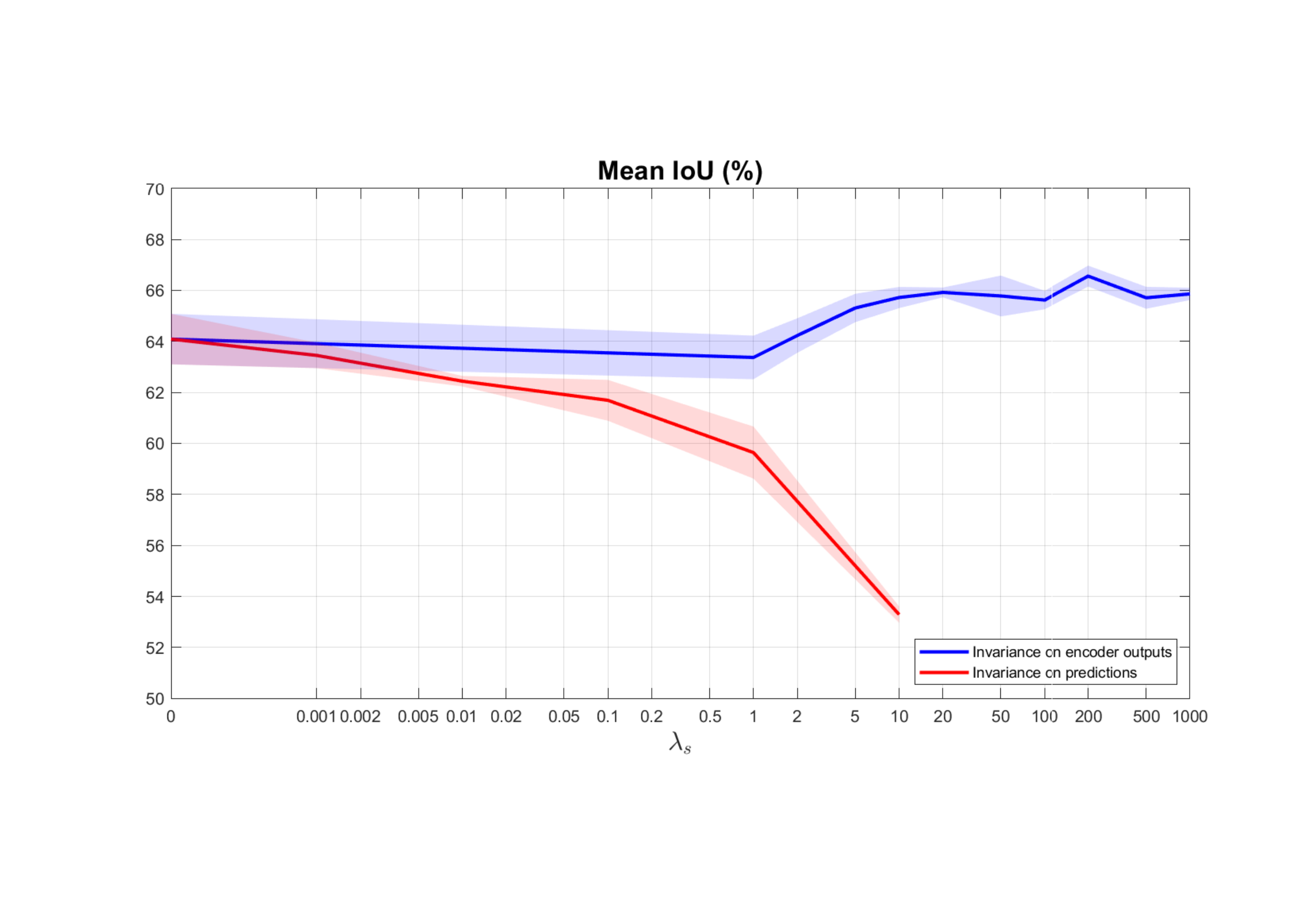}
    \caption{\textbf{Ablation of the point in the network where invariance is applied on Cityscapes$\to$ACDC.} Evaluation is performed on the validation set of ACDC. The $x$-axis is logarithmic and shows the weight $\lambda_s$ of the feature invariance loss, which is applied here only on the source domain. Averages and standard deviations are plotted over three runs for each configuration. The two plotted lines share their leftmost point, which corresponds to $\lambda_s = 0$, i.e., not applying an invariance loss at all.}
    \label{fig:ablation:invariance:point}
\end{figure}

\begin{table}[!tb]
  \caption{\textbf{Hyperparameter study of the weights of our feature invariance losses on Cityscapes$\to$ACDC.} Evaluation is performed on the validation set of ACDC. Mean and standard deviation of mean IoU across three runs are reported.}
  \label{table:ablation:weights}
  \centering
  \renewcommand{\arraystretch}{1.1}
  \setlength\tabcolsep{2pt}
  \footnotesize
  \smallskip
  \begin{tabular*}{\linewidth}{l @{\extracolsep{\fill}} ccccc}
  \toprule
  $\lambda_s$ & 50 & 100 & 200 & 500 & 1000\\
  \midrule
  CISS-source & 65.8$\pm$1.6 & 65.6$\pm$0.8 & \textbf{66.6}$\pm$0.8 & 65.7$\pm$0.9 & 65.9$\pm$0.5\\
  \midrule
  $\lambda_t$ & 20 & 50 & 100 & 200 & 500\\
  \midrule
  CISS-target & 66.7$\pm$0.6 & 66.6$\pm$1.6 & \textbf{66.9}$\pm$0.5 & 66.1$\pm$0.7 & 65.2$\pm$0.6\\
  \bottomrule
  \end{tabular*}
\end{table}

\begin{figure}
    \centering
    \includegraphics[width=\linewidth,trim={0.7cm 8.3cm 0.1cm 7.0cm},clip]{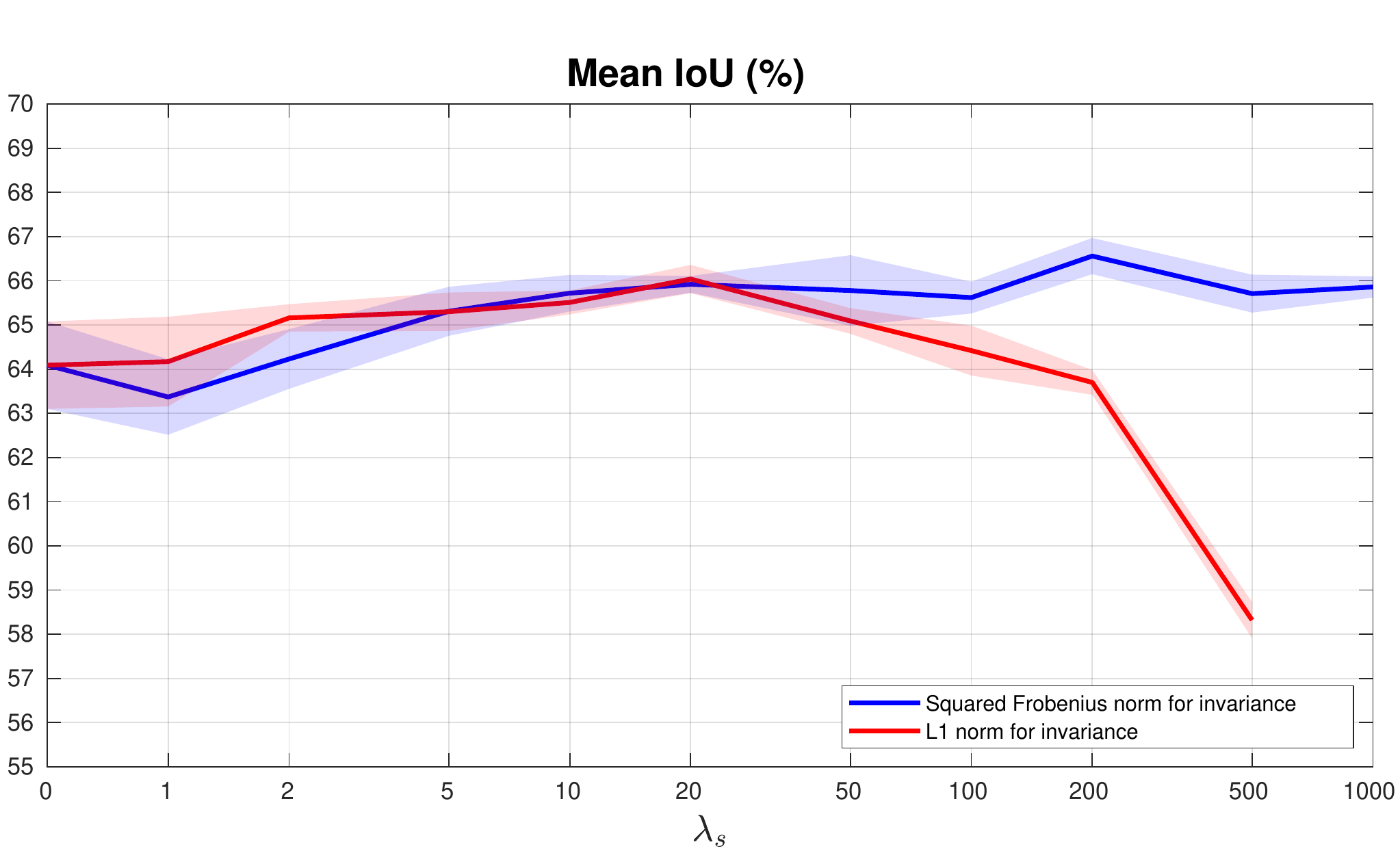}
    \caption{\textbf{Ablation of the norm which is used in the feature invariance loss on Cityscapes$\to$ACDC.} Evaluation is performed on the validation set of ACDC. The $x$-axis is logarithmic and shows the weight $\lambda_s$ of the feature invariance loss, which is applied here only on the source domain. Averages and standard deviations are plotted over three runs for each configuration. Results with the proposed, squared Frobenius norm are plotted in blue and those with the alternative, $L_1$ norm are plotted in red. The two plotted lines share their leftmost point, which corresponds to $\lambda_s = 0$, i.e., not applying an invariance loss at all.}
    \label{fig:ablation:invariance:norm}
\end{figure}

\begin{figure}
    \centering
    \subfloat[Losses for baseline model with $\mathcal{L}_{\text{basic}}$ from \eqref{eq:loss:basic}]{\includegraphics[width=\linewidth,trim={4.2cm 8.5cm 4.5cm 8.0cm},clip]{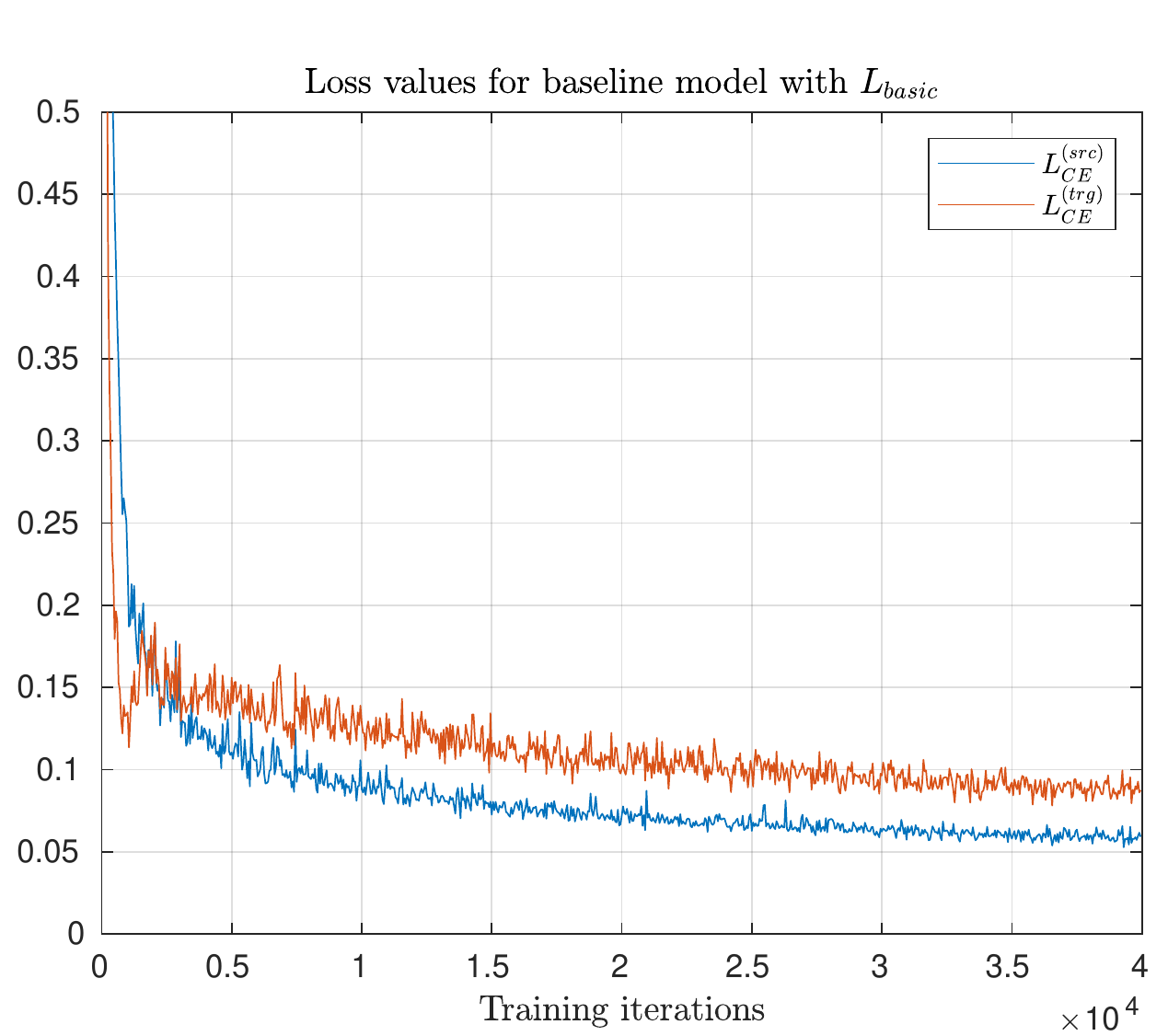}\label{fig:ablation:invariance:convergence:basic}}
    \hfil
    \subfloat[Losses for CISS model with $\mathcal{L}_{\text{CISS}}$ from \eqref{eq:loss:ciss}]{\includegraphics[width=\linewidth,trim={4.2cm 8.5cm 4.5cm 8.0cm},clip]{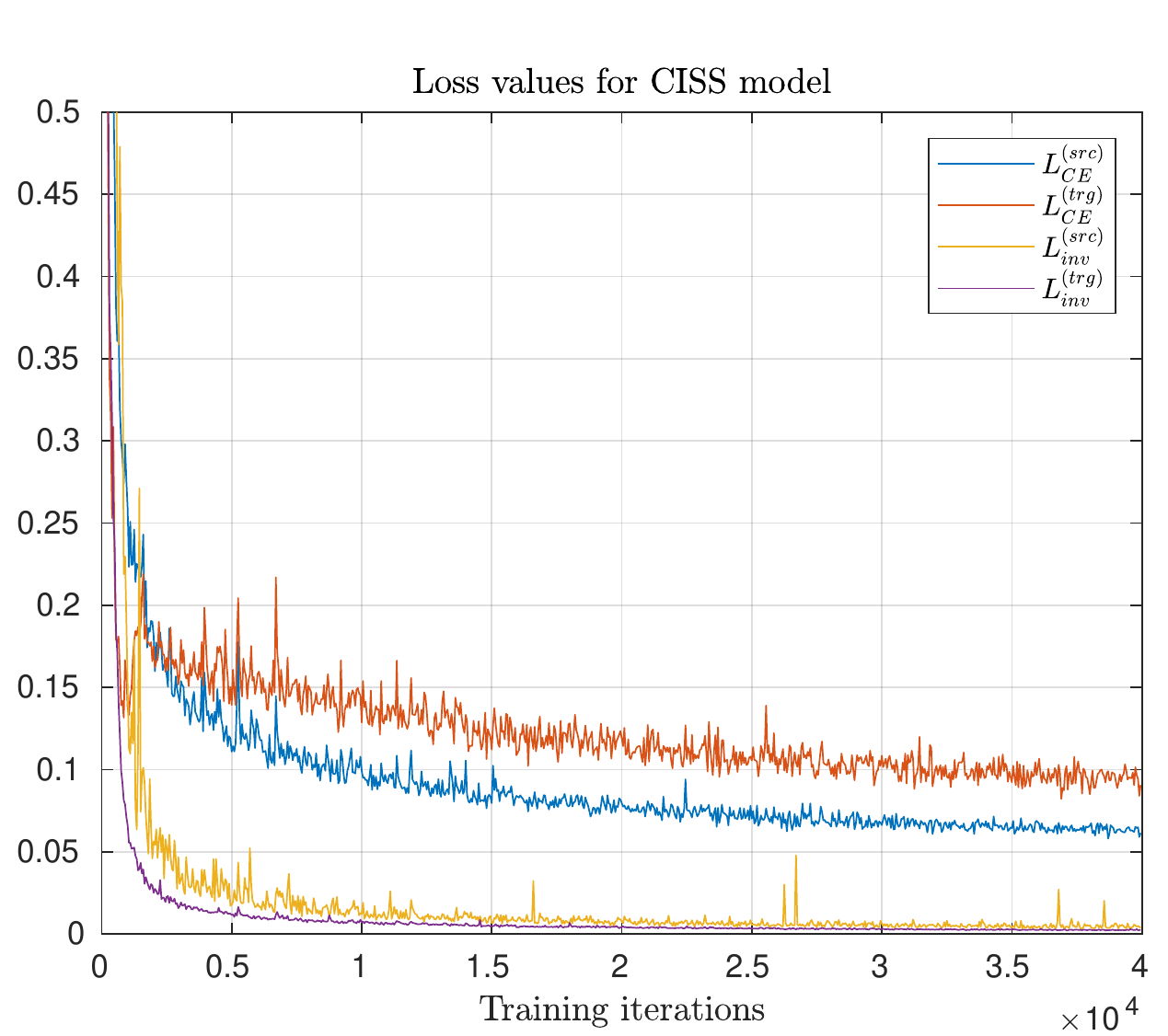}\label{fig:ablation:invariance:convergence:ciss}}
    \caption{\textbf{Evolution and convergence of training losses on Cityscapes$\to$Dark Zurich.} Loss evaluation is performed on training samples from the training sets of Cityscapes and Dark Zurich. In \protect\subref{fig:ablation:invariance:convergence:basic}, the curves for the two cross-entropy losses (one on the source domain and the other on the target domain) of the baseline, HRDA-equivalent model from \eqref{eq:loss:basic} are shown in blue and red, to provide broader context. In \protect\subref{fig:ablation:invariance:convergence:ciss}, we show the training loss curves for our proposed CISS model from \eqref{eq:loss:ciss}, both for the two aforementioned cross-entropy losses (in blue and red) and additionally for the two feature invariance losses $\mathcal{L}_{\text{inv}}$ (in yellow and purple) which are involved in our training.}
    \label{fig:ablation:invariance:convergence}
\end{figure}

\subsubsection{Effect of Weights of Feature Invariance Losses}
\label{sec:experiments:ablations:weights}

We examine the influence of the value of the two hyperparameters of CISS, i.e., the weights $\lambda_s$ and $\lambda_t$ of the two feature invariance losses, on performance in Table~\ref{table:ablation:weights}. In particular, we consider the ablated versions of CISS in which either of the two feature invariance losses is included, the source one (CISS-source) or the target one (CISS-target), and vary the respective weight. The best performance is obtained at $\lambda_s = 200$ for CISS-source and at $\lambda_t = 100$ for CISS-target. However, note that performance degrades gracefully as we move away from these values, implying that our method is fairly insensitive to the exact values of these hyperparameters.

\subsubsection{Benefit of Internal Feature Invariance Loss Versus Output Invariance Loss}
\label{sec:experiments:ablations:internal:vs:output}

We justify the choice of applying feature invariance to the encoder outputs, i.e., the \emph{internal} features of the network, via the experiment of Fig.~\ref{fig:ablation:invariance:point}. The result shown in Fig.~\ref{fig:ablation:invariance:point} verifies our intuition that invariance on internal features works better than on network outputs. In particular, using the invariance loss in the source domain, its application to internal features can improve significantly upon not using the invariance loss at all when the respective weight $\lambda_s$ is tuned properly, while its application to network outputs, i.e.\ predictions, invariably deteriorates performance compared to not enforcing invariance at all.

\subsubsection{Study of Norm Used in Feature Invariance Loss}
\label{sec:experiments:ablations:norm}

\blue{The default formulation of our novel feature invariance loss in the framework of CISS in \eqref{eq:feature:invariance} involves the squared Frobenius norm of the difference between the feature tensors associated with different views. The rationale of applying this $L_2$-like loss, as opposed to robust losses such as $L_1$ or Huber, is that we aim at assigning a larger penalty to very large deviations in corresponding internal features, even when such deviations only occur at few spatial locations. In other words, we aim to impose feature invariance \emph{everywhere} in the input images, which is achieved better with the proposed Frobenius norm of \eqref{eq:feature:invariance}. Sparse large deviations in the two corresponding feature maps, which result from robust losses such as $L_1$ or Huber, are not desirable, as invariance should hold globally. More formally, an alternative, $L_1$-norm-based formulation of our feature invariance loss is
\begin{multline}
    \label{eq:feature:invariance:l1}
    \mathcal{L}_{\text{inv},L_1}(F=\omega\circ\phi, I, I^{\prime})\\
    = \frac{1}{DMN}\sum_{d=1}^{D}\sum_{m=1}^{M}\sum_{n=1}^{N}\left|(\phi(I))_{dmn} - (\phi(I^{\prime}))_{dmn}\right|.
\end{multline}
We compare this $L_1$-based instantiation of the feature invariance loss to the proposed squared Frobenius instantiation of \eqref{eq:feature:invariance} in Fig.~\ref{fig:ablation:invariance:norm}, by considering the ablated CISS-source version of our method for simplicity. In particular, we train both variants---the one based on \eqref{eq:feature:invariance} and the one based on \eqref{eq:feature:invariance:l1}---for varying values of the loss weight $\lambda_s$. We observe that our proposed squared Frobenius norm achieves better peak performance overall and is more robust to the precise choice of the loss weight $\lambda_s$ than the $L_1$ norm. Still, the $L_1$-based feature invariance loss from \eqref{eq:feature:invariance:l1} also improves over the baseline from \eqref{eq:loss:basic}, which does not use an invariance loss at all (leftmost data point in Fig.~\ref{fig:ablation:invariance:norm}), across a wide range of values of $\lambda_s$, i.e.\ for $\lambda_s \in [1,100]$. This implies that even though CISS works best when using our proposed squared Frobenius norm for penalizing differences between internal features, it is not strictly specific to this formulation and it works decently well with other instantiations of the feature invariance loss too.}

\subsubsection{Convergence of Feature Invariance Losses}
\label{sec:experiments:ablations:convergence}

\blue{Although our proposed optimization objective in \eqref{eq:loss:ciss} encourages the minimization of deviations between internal features of different views of the same scene via our feature invariance loss, it is necessary to examine to what extent such deviations are actually minimized towards the end of the training process as well as whether and how this minimization affects the concurrent minimization of the basic, cross-entropy losses for semantic segmentation which are also involved in \eqref{eq:loss:ciss}. We perform this analysis in Fig.~\ref{fig:ablation:invariance:convergence}, in which we examine the evolution of the above training losses in the Cityscapes$\to$Dark Zurich setting. First, we observe that the two basic cross-entropy losses on the source domain and the target domain evolve in a very similar way both in the case where no feature invariance loss is applied (Fig.~\ref{fig:ablation:invariance:convergence:basic}) and in the case where our fully-fledged CISS framework with feature invariance losses is used (Fig.~\ref{fig:ablation:invariance:convergence:ciss}). This evidences the harmlessness of our feature invariance loss for the simultaneous optimization of the main semantic segmentation objectives in the examined domain adaptation setting. What is more, we observe in Fig.~\ref{fig:ablation:invariance:convergence:ciss} that our feature invariance losses in CISS both for the source domain (yellow) and the target domain (purple) converge very well to $0$, implying that the goal of feature invariance across the source and target domain is achieved effectively with CISS.}

\subsubsection{Generality of CISS with Respect to Stylization Method}
\label{sec:experiments:ablations:stylization}

\begin{table}[!tb]
  \caption{\textbf{Comparison of CISS using different stylization techniques for applying feature invariance loss in the target domain for Cityscapes$\to$ACDC adaptation.} Evaluation is performed on the validation set of ACDC. We compare the default FDA~\cite{fda:adaptation} stylization, the stylization using the method by Reinhard et al.~\cite{reinhard2001color}, and the simple color jitter augmentation which is detailed in Sec.~\ref{sec:experiments:setup}. Mean and standard deviation across three runs are reported.}
  \label{table:ablation:stylization}
  \centering
  \setlength\tabcolsep{2pt}
  \footnotesize
  \begin{tabular}{lc}
  \toprule
  Invariance Loss & Mean IoU (\%)\\
  \midrule
  None & 64.1$\pm$2.0\\
  With FDA stylization ($\lambda_t = 100$) & 66.9$\pm$0.5\\
  With Reinhard stylization ($\lambda_t = 2$) & 66.7$\pm$0.7\\
  With color jitter augmentation ($\lambda_t = 50$) & 67.3$\pm$1.0\\
  \bottomrule
  \end{tabular}
\end{table}

We test CISS in Table~\ref{table:ablation:stylization} with the color transfer technique in \cite{reinhard2001color} \blue{as well as with simple color jitter augmentation} for stylizing \blue{resp.\ augmenting} the input images, in order to verify the generality of CISS with regard to the stylization \blue{or augmentation} method \blue{$g$ from \eqref{eq:stylization:source} and \eqref{eq:stylization:target}} that is used for imposing feature invariance. In particular, we consider the case where feature invariance is applied in the target domain and test CISS (i) with the default FDA stylization, (ii) with \cite{reinhard2001color}, \blue{and (iii) with color jitter augmentation on the target-domain images}, setting the optimal weight $\lambda_t$ separately for each variant. CISS improves significantly \blue{in all cases} upon the baseline that does not use any feature invariance and it achieves similar performance with \blue{all three} stylization\blue{/augmentation} methods, which evidences the generality of CISS with regard to the method that it employs for altering the appearance/style of the input images. \blue{In particular, for the color jitter augmentation case, CISS even performs slightly better than with FDA, indicating that CISS is indeed not specific to or reliant on FDA.}

\section{Conclusion}
\label{sec:conclusion}

We have presented CISS, a UDA method for semantic segmentation tailored for condition-level domain shifts. Our method promotes invariance of the internal features that are extracted by the semantic segmentation network to visual conditions, which are modeled through the style of the input, by penalizing the difference between features of the same image when the latter is rendered in the styles of the source and the target domain. We have performed a thorough experimental evaluation of CISS and showed that it excels on normal-to-adverse condition-level adaptation from Cityscapes to Dark Zurich and from Cityscapes to ACDC. Our model which has been adapted from Cityscapes to Dark Zurich generalizes much better to other unseen nighttime domains, \blue{such as} BDD100K-night \blue{and ACDC-night}, than competing state-of-the-art models, demonstrating that condition invariance makes models trained with CISS more robust to diverse inputs. Last but not least, we have shown that the novel feature-level alignment performed by CISS on internal features works (i) much better than output-level alignment, and (ii) irrespective of the particular stylization method that CISS employs.

\ifCLASSOPTIONcompsoc
  \section*{Acknowledgments}
\else
  \section*{Acknowledgment}
\fi

This work is funded by Toyota Motor Europe via the research project TRACE-Z\"urich.

\bibliographystyle{IEEEtran}
\bibliography{IEEEabrv,refs}

\begin{IEEEbiography}[{\includegraphics[width=1in,clip,keepaspectratio]{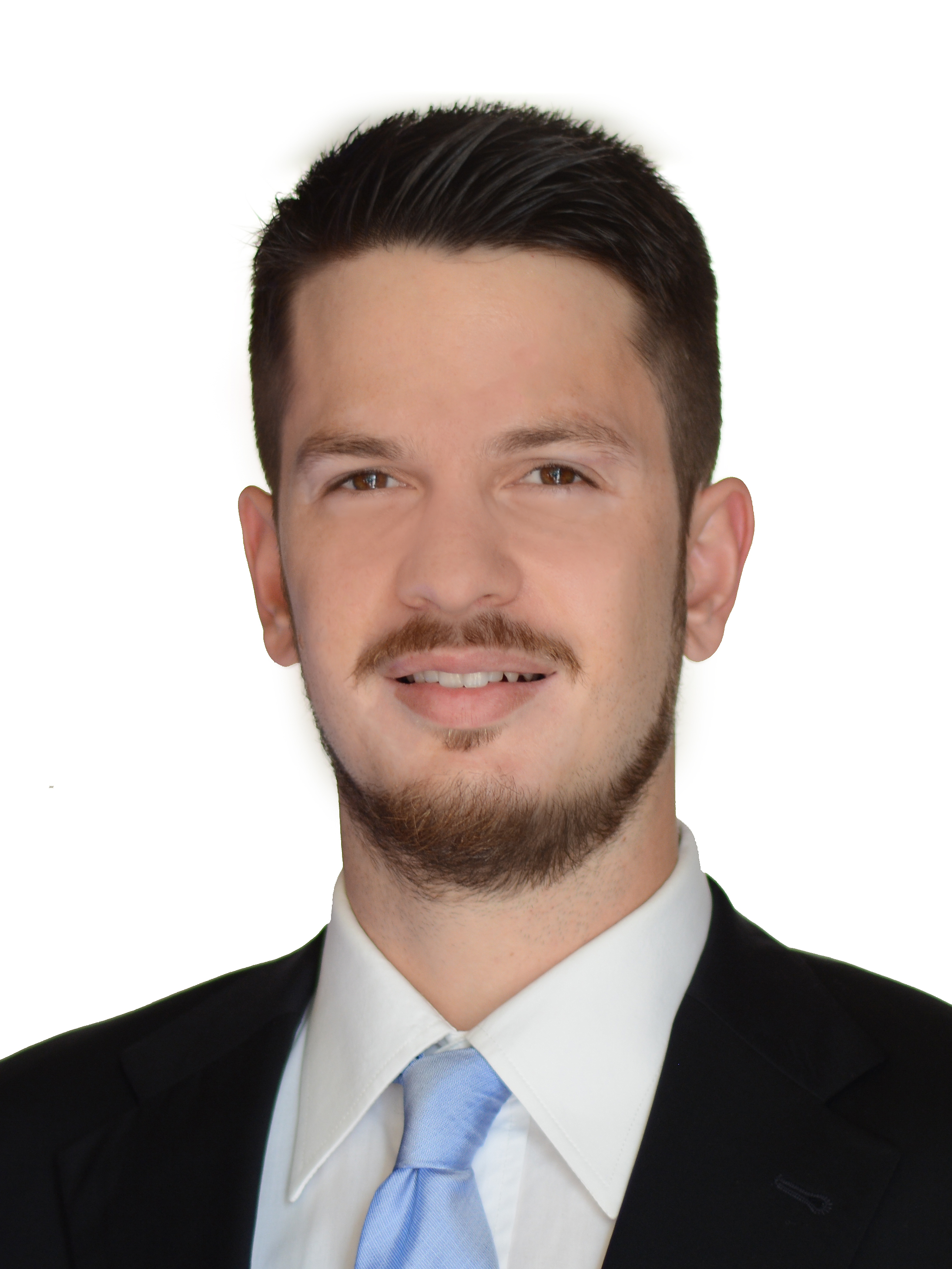}}]{Christos Sakaridis}
is a lecturer at ETH Zurich and a senior postdoctoral researcher at Computer Vision Lab of ETH Zurich. His research fields are computer vision and machine learning. The focus of his research is on semantic and geometric visual perception, involving multiple domains, visual conditions, and visual or non-visual modalities. Since 2021, he is the Principal Engineer in TRACE-Zurich, a large-scale project on computer vision for autonomous cars and robots. He received the ETH Zurich Career Seed Award in 2022. He obtained his PhD from ETH Zurich in 2021, having worked in Computer Vision Lab. Prior to that, he received his MSc in Computer Science from ETH Zurich in 2016 and his Diploma in Electrical and Computer Engineering from National Technical University of Athens in 2014.
\end{IEEEbiography}

\begin{IEEEbiography}[{\includegraphics[width=1in,clip,keepaspectratio]{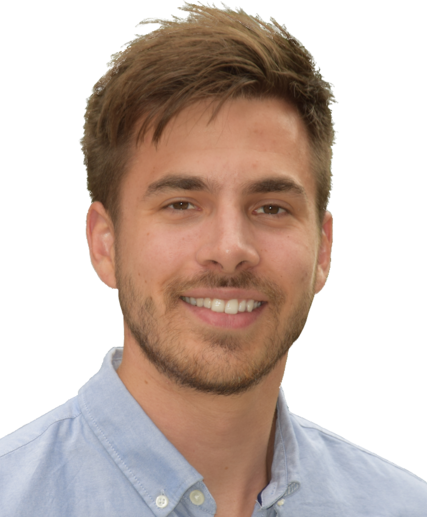}}]{David Bruggemann}
is a doctoral candidate at Computer Vision Lab, ETH Zurich. His research focuses on designing neural networks which can learn multiple visual tasks concurrently and are robust to changing visual conditions. To this end, he also explores alternative input modalities, such as lidars, event cameras, and radars. Prior to joining Computer Vision Lab, he received his BSc and MSc degrees in Mechanical Engineering from ETH Zurich in 2016 and 2019, respectively.
\end{IEEEbiography}

\vfill

\newpage

\begin{IEEEbiography}[{\includegraphics[width=1in,clip,keepaspectratio]{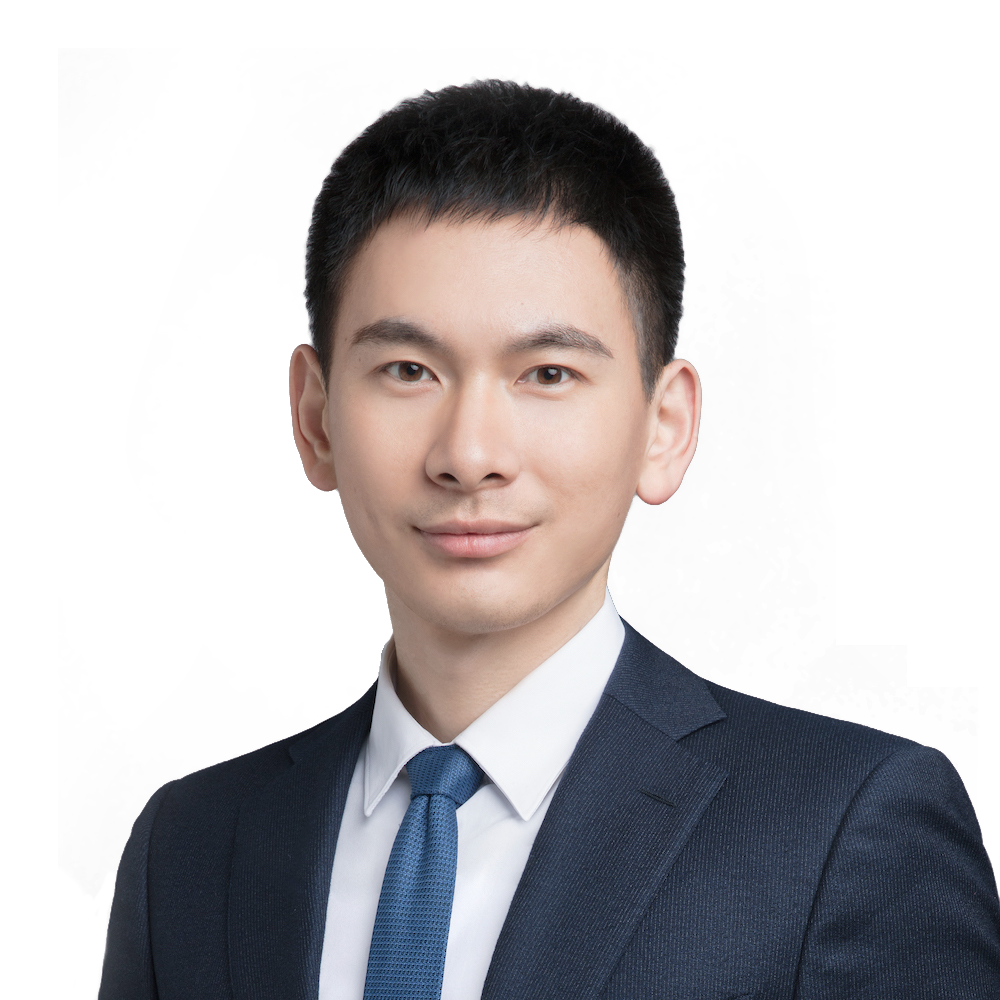}}]{Fisher Yu}
is an assistant professor at ETH Zurich. He obtained his PhD from Princeton University and became a postdoctoral researcher at UC Berkeley afterwards. He directs the Visual Intelligence and Systems Group at Computer Vision Lab, ETH Zurich. His goal is to build perceptual systems capable of performing complex tasks in complex environments. His research is at the junction of machine learning, computer vision, and robotics. He currently works on closing the loop between vision and action.
\end{IEEEbiography}

\begin{IEEEbiography}[{\includegraphics[width=1in,clip,keepaspectratio]{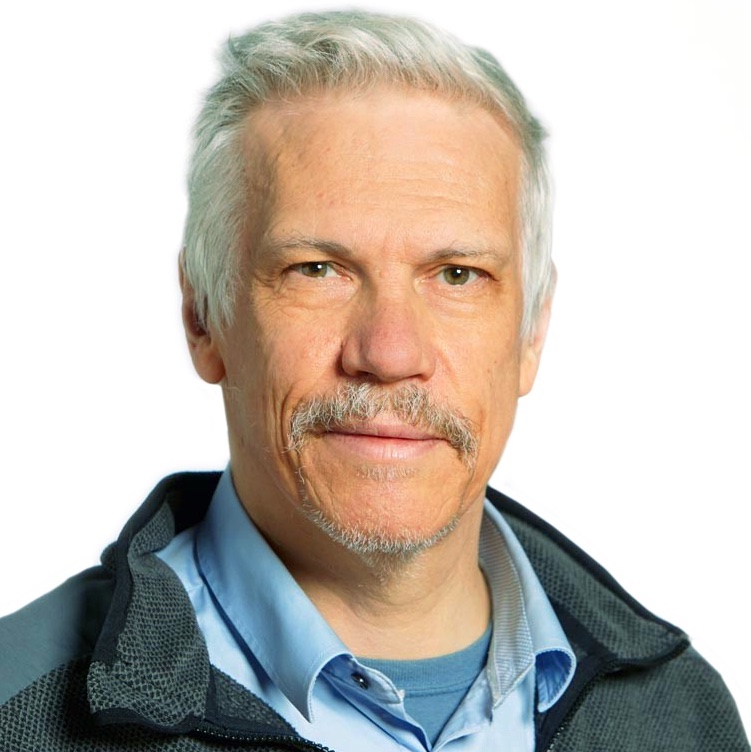}}]{Luc Van Gool}
is a full professor for Computer Vision at ETH Zurich, the KU Leuven and INSAIT. He leads research and/or teaches at all three institutions. He has authored over 900 papers. He has been a program committee member of several major computer vision conferences (e.g.\ Program Chair ICCV’05, Beijing, General Chair of ICCV’11, Barcelona, and of ECCV’14, Zurich). His main interests include 3D reconstruction and modeling, object recognition, and autonomous driving. He received several best paper awards (e.g.\ David Marr Prize ’98, Best Paper CVPR’07). He received the Koenderink Award in 2016 and the ``Distinguished Researcher'' nomination by the IEEE Computer Society in 2017. In 2015 he also received the 5-yearly Excellence Prize by the Flemish Fund for Scientific Research. He was the holder of an ERC Advanced Grant (VarCity). Currently, he leads computer vision research for autonomous driving in the context of the Toyota TRACE labs at ETH and in Leuven, and has an extensive collaboration with Huawei on the topic of image and video enhancement.
\end{IEEEbiography}

\vfill

\end{document}